\documentclass[runningheads]{llncs}

\usepackage{eccv}

\usepackage{eccvabbrv}

\usepackage{graphicx}
\usepackage{booktabs}
\usepackage{adjustbox}

\usepackage[accsupp]{axessibility}  

\usepackage{hyperref}

\usepackage{orcidlink}

\usepackage{amsmath,amsfonts,bm}

\def\eqref#1{equation~\ref{#1}}

\def\1{\bm{1}}

\def\vf{{\bm{f}}}

\def\vx{{\bm{x}}}

\def\mS{{\bm{S}}}

\def\mW{{\bm{W}}}
\def\mX{{\bm{X}}}

\DeclareMathAlphabet{\mathsfit}{\encodingdefault}{\sfdefault}{m}{sl}
\SetMathAlphabet{\mathsfit}{bold}{\encodingdefault}{\sfdefault}{bx}{n}

\def\sR{{\mathbb{R}}}

\usepackage{multirow}
\usepackage[dvipsnames, svgnames, table]{xcolor}
\usepackage{wrapfig}
\usepackage{fontawesome}
\usepackage{marvosym}

\newcommand{\detectModel}{\mathcal{D}}
\newcommand{\reidModel}{\Phi}

\newcommand{\bboxWithti}[2]{\mathbf{b}_{#1, #2}}

\newcommand{\feat}{\mathbf{f}}
\newcommand{\featWithti}[2]{\mathbf{f}_{#1, #2}}

\newcommand\blfootnotetext[1]{%
  \begingroup
  \renewcommand{\thefootnote}{}%
  \footnotetext{#1}%
  \endgroup
}

\newcommand{\featEMAWithtj}[2]{\hat{\mathbf{f}}_{#1, \tau_{#2}}}

\newcommand{\improve}[1]{\textcolor{ForestGreen}{\tiny (#1)}}
\newcommand{\drop}[1]{\textcolor{Crimson}{\tiny (#1)}}

\newcommand{\citep}{\cite}
\newcommand{\citet}{\cite}

\begin{document}

\title{History-Aware Transformation of ReID Features for Multiple Object Tracking} 

\titlerunning{HATReID-MOT}

\author{Ruopeng Gao\inst{1}\orcidlink{0009-0002-5522-5488} \and
Yuyao Wang\inst{1}\orcidlink{0009-0002-6759-6569} \and
Chunxu Liu\inst{1}\orcidlink{0009-0005-5179-4116} \and
Limin Wang\inst{1,2,}\textsuperscript{\footnotesize \Letter}\orcidlink{0000-0002-3674-7718}}


\institute{Nanjing University, Nanjing, China \and
Shanghai AI Laboratory, Shanghai, China \\
\url{https://github.com/MCG-NJU/HATReID-MOT} \\
\email{\{ruopenggao,wayfareryy\}@gmail.com \quad chunxu.liu@smail.nju.edu.cn \quad lmwang@nju.edu.cn}}

\maketitle

\blfootnotetext{\Letter~: Corresponding author.}

\begin{abstract}
    In Multiple Object Tracking (MOT), Re-identification (ReID) features are widely employed as a powerful cue for object association. 
    However, they are often wielded as a one-size-fits-all hammer, applied uniformly across all videos through simple similarity metrics. We argue that this overlooks a fundamental truth: MOT is not a general retrieval problem, but a context-specific task of discriminating targets within a single video.
    To this end, we advocate for the adjustment of visual features based on the context specific to each video sequence for better adaptation.
    In this paper, we propose a history-aware feature transformation method that dynamically crafts a more discriminative subspace tailored to each video's unique sample distribution. Specifically, we treat the historical features of established trajectories as context and employ a tailored Fisher Linear Discriminant (FLD) to project the raw ReID features into a sequence-specific representation space.
    Extensive experiments demonstrate that our training-free method dramatically enhances the discriminative power of features from diverse ReID backbones, resulting in marked and consistent gains in tracking accuracy.
    Our findings provide compelling evidence that MOT inherently favors context-specific representation over the direct application of generic ReID features.
    We hope our work inspires the community to move beyond the naive application of ReID features and towards a deeper exploration of their purposeful customization for MOT.
    \keywords{Multi-object tracking \and MOT \and Tracking}
\end{abstract}

\section{Introduction}
\label{sec:introduction}

Multiple Object Tracking (MOT) is a fundamental computer vision task that aims to detect objects and maintain their identities across video frames. Its primary goal is to generate a distinct trajectory for each target by associating its corresponding detections over time \cite{MIR-MOT}. As a critical component for understanding dynamic scenes, MOT serves as an essential prerequisite for a wide range of downstream applications, such as autonomous driving \cite{MIR-AutoDrive}, human behavior analysis \cite{Unified-MOT-Action}, trajectory forecasting \cite{Joint-MOT-Trajectory-Prediction}, and public surveillance.

The tracking-by-detection paradigm \citep{SORT, ByteTrack, OC-SORT} has long been the dominant and most widely adopted approach in the field of multiple object tracking. 
According to the task definition, it decouples the complex tracking problem into two sequential subtasks: first, an object detector localizes all targets within each frame, and second, an association algorithm links these detections across frames to form individual trajectories.
As the former step is well-addressed by powerful detectors \citep{YOLOX, YOLOv8}, the crux of this paradigm lies in the association stage.
To solve this association problem, most methods \citep{FairMOT, OC-SORT, QuoVadis} model existing trajectories with discriminative cues and then allocate identities by minimizing the matching cost.

\begin{figure}[tb]
  \centering
  \begin{subfigure}{0.54\linewidth}
    \includegraphics[width=0.96\linewidth]{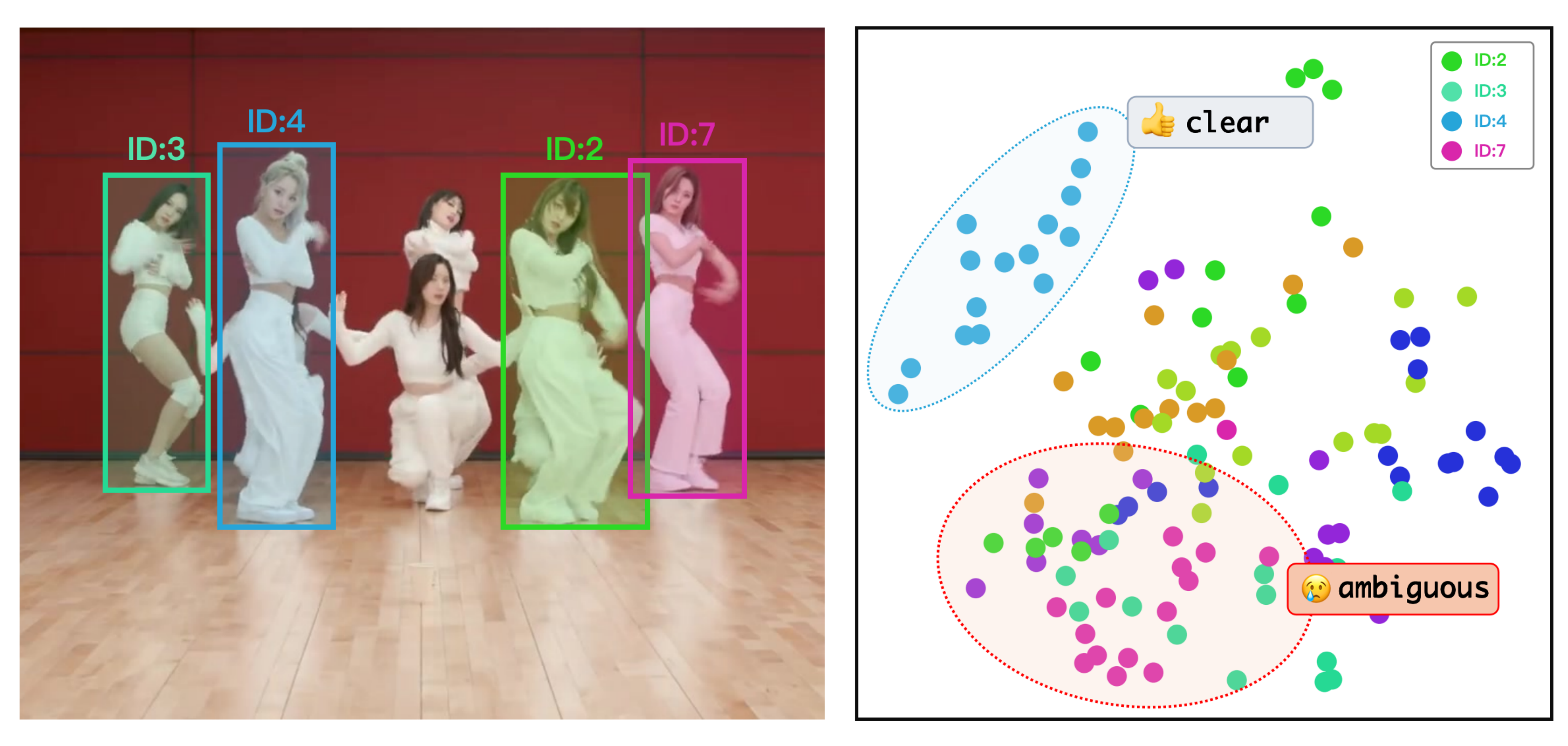}
    \caption{High similarity between the ReID features of different trajectories can create potential ambiguity.}
    \label{fig:space-one-sequence}
  \end{subfigure}
  \hfill
  \begin{subfigure}{0.44\linewidth}
    \includegraphics[width=0.95\linewidth]{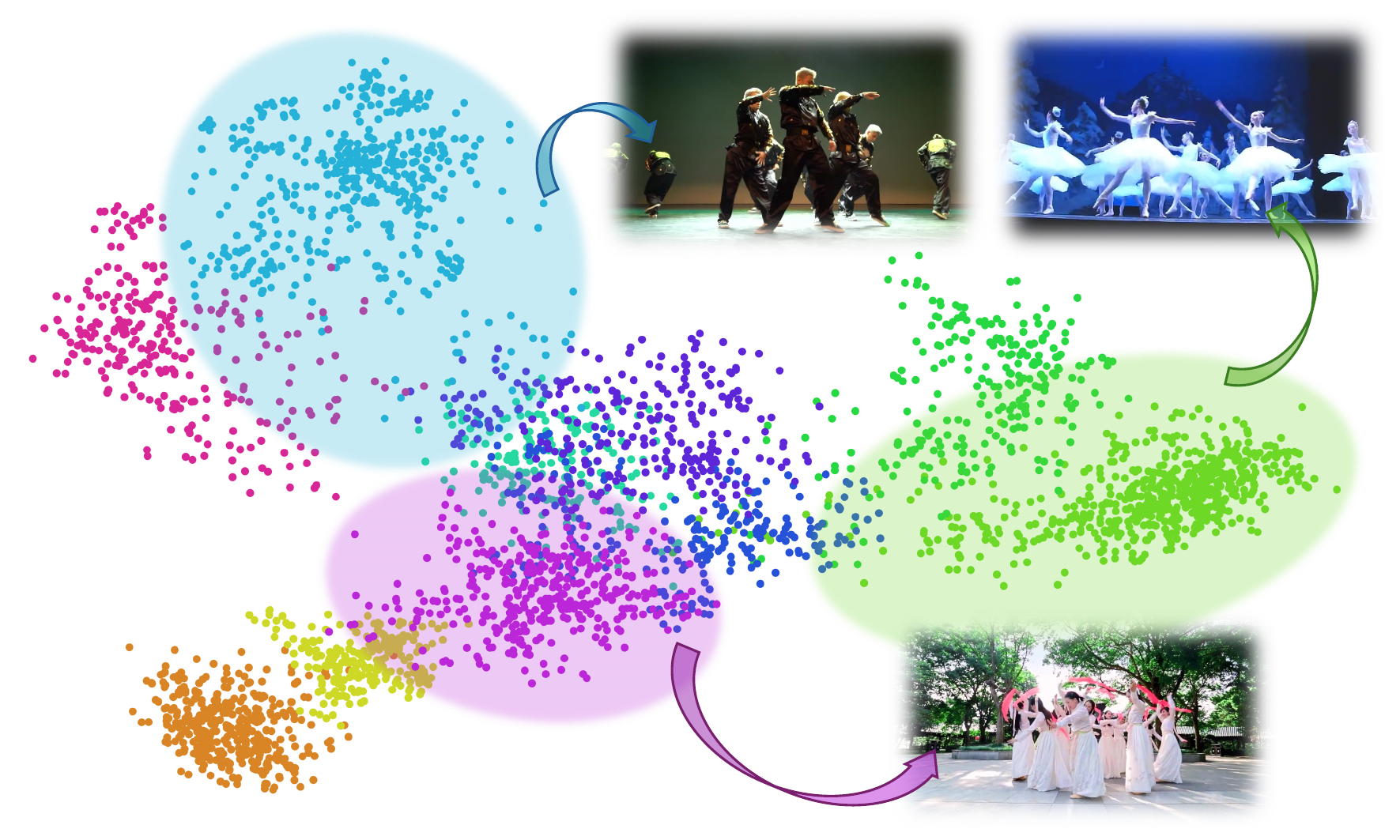}
    \caption{ReID features of individual sequences form compact clusters in the original space.}
    \label{fig:space-all-sequence}
  \end{subfigure}
  \caption{Visualization of ReID features in the original representation space \citep{FastReID}.}
  \label{fig:sequence}
\end{figure}

Given that distinct targets often exhibit unique visual characteristics, appearance has emerged as a powerful and prevalent discriminative feature for trajectory modeling.
In practice, visual features are typically extracted using off-the-shelf Re-Identification (ReID) models \citep{FastReID}, and a cost matrix is then formulated by the cosine distance.
Despite its demonstrated success \citep{Deep-SORT, FairMOT, BoT-SORT, Hybrid-SORT}, a latent contradiction persists within this paradigm.
According to the definition, the goal of a general ReID model is to learn a universal feature representation capable of distinguishing any given identity from a large, open set. In contrast, the challenge within MOT is to discriminate only between the limited set of targets appearing in a specific video, which constitutes a more nuanced, expert-level requirement.
As illustrated in \cref{fig:space-one-sequence}, targets within the same video sequence always exhibit a high degree of similarity, making them difficult to distinguish in a generic, globally-trained feature space \citep{FastReID}. Furthermore, this intra-sequence similarity causes their representations to cluster within a confined subspace of the original space, leading to redundancy, as shown in \cref{fig:space-all-sequence}.
Based on the foregoing observations, a natural question arises: \textit{\textbf{can we seek a specialized representation subspace for MOT, one that is more focused on distinguishing identities within the constrained set of a given sequence?}}

In this paper, we first confirm the significant influence of representation discriminability on tracking performance. Accordingly, we posit that an ideal representation space should be structured for each sequence to minimize intra-trajectory distances while maximizing inter-trajectory distances. Such a configuration would effectively increase the separability between positive and negative pairs, thereby enhancing the tracking ability.
Coincidentally, this principle is conceptually analogous to the objective of Fisher Linear Discriminant (FLD) \citep{FLD}, where trajectories are viewed as different categories.
Since tracking is an online, continuous inference process, the established historical context can be regarded as a dynamic approximation of the data distribution for the current sequence.
Building on these discussions, in practice, we feed the existing trajectories as conditional input into the FLD algorithm. By computing the closed-form solution, we derive a projection matrix that maps the original features onto a more discriminative subspace tailored for separating different trajectories.
The experiment results reveal that this simple yet effective feature transformation substantially enhances discriminability and boosts overall tracking performance.
Nevertheless, we revisit this process and argue that MOT possesses some task-specific demands that should not be overlooked.
Firstly, since a target's features gradually change over time in online tracking, we use a temporally weighted average to construct the trajectory's center, rather than the na\"ive averaging. This makes the resulting representation better suited for the similarity assessment required at the current timestep.
Secondly, historical trajectories are not always reliable due to occlusions and tracking errors. Moreover, the tracker needs to handle newborn objects, which are not considered in the transformed space. These challenges underscore the importance of retaining the original feature space with its strong robustness and generalization capabilities.
To this end, we combine the similarity scores from both the general and specialized representations, thereby leveraging the complementary strengths of each.

To clearly validate the impact of our ReID feature transformation, our experiments are primarily conducted on trackers that rely solely on appearance \citep{FastReID, MASA}. This approach minimizes the complex designs and potential interference introduced by other tracking cues \citep{KF, SORT}.
In practice, we build a ReID-based tracker upon the most widely-used ReID model \citep{FastReID} in the MOT community \citep{Hybrid-SORT, DiffMOT} and validate the effectiveness of our components. Relying solely on the ReID cue, our method achieves significant performance improvements.
Remarkably, in some scenarios \citep{SportsMOT}, our algorithm substantially outperforms methods that combine multiple clues \citep{BoT-SORT, SportsMOT, DiffMOT}, establishing a new state-of-the-art result. This finding strongly indicates that the full potential of appearance information has been underestimated in past research.
We also confirm the generalization capability of our proposed method by applying it to \citet{MASA} with various visual encoders \citep{ResNet, Detic, SAM, Grounding-DINO}, observing stable performance boosts across every case.
Additionally, we conduct experiments on several hybrid-based methods \citep{OC-SORT, Hybrid-SORT}. The results demonstrate that our approach can be seamlessly integrated into these advanced trackers, achieving state-of-the-art performance.



To sum up, our main contributions include:

\begin{itemize}
    \item Following our analysis in \cref{sec:preliminary-analysis}, we equip Fisher Linear Discriminant with historical tracklet supervision to transform ReID features, enhancing their discriminability. 
    \item To address the practical needs of MOT task, we propose two customized components, \textit{temporally-weighted trajectory centroid} (\cref{sec:method-temporal}) and \textit{knowledge integration} (\cref{sec:method-integration}), which further improve our tracking performance.
    \item To prove the effectiveness of our method, we conduct extensive experiments on ReID-based methods, demonstrating consistent performance gains across diverse scenarios (\cref{tab:sota-dancetrack}, \cref{tab:sota-sportsmot} and \cref{tab:sota-masa}). We also validate its versatility by seamlessly integrating it into hybrid-based trackers, pushing their state-of-the-art performance even further.
\end{itemize}
\section{Related Work}
\label{sec:related}

\textbf{Tracking-by-Detection} methods decouple the multiple object tracking (MOT) task into two sub-tasks: object detection and data association.
While a minority of studies \citep{detecting-invisible} have explored customized detection methods, the vast majority of research \citep{CenterTrack, ByteTrack, TrackFlow, SparseTrack, AgriSORT} has focused on the design of the target association algorithm.
In this process, researchers model trajectories and measure affinities based on diverse cues.
For visual appearance, most methods \citep{BoT-SORT, Deep-OC-SORT, Hybrid-SORT, DiffMOT} directly utilize off-the-shelf ReID models \citep{FastReID} to extract features. While some approaches \citep{FairMOT, JDE, ConstrastiveDETR-MOT} employ custom-designed extractors, they still adhere to the fundamental principles and supervision methods of traditional ReID methods \citep{ReID, FastReID}.
Regarding location information, the most classic method \citep{SORT, ByteTrack} is to use the Kalman filter \citep{KF} for linear estimation of the motion. To handle non-linear dynamics \citep{DanceTrack, SportsMOT} and other complex cases, recent methods have introduced many tailored rules \citep{OC-SORT, StrongSORT, Hybrid-SORT, UCMCTrack} or adopted learnable modules for motion prediction \citep{QuoVadis, MotionTrack, DiffusionTrack, DiffMOT, MambaTrack, MambaTrack+}.
Many approaches \citep{Deep-SORT, StrongSORT, Deep-OC-SORT, Hybrid-SORT, DiffMOT} also fuse the two aforementioned cues together to fully leverage their respective advantages.
Furthermore, some other methods introduce even more information modalities, such as Bird's-Eye-View (BEV) perspectives \citep{QuoVadis} and depth information \citep{BoT-SORT, TrackFlow, PD-SORT}.
Most relevant to our work are several studies that aim to customize the ReID branch of the MOT task: \citet{Temporal-Global-Local-ReID-MOT} seeks to mitigate the mismatch between its global temporal training and local temporal inference, \citet{MOT-long-tail} performs group-wise similarity calculation to address the long-tail distribution problem, \citet{Single-Shot-ReID-MOT} helps newborn targets acquire more robust representations, \citet{TOPICTrack} sharpens the distinction in similarity, \citet{CATrack} employs a condition-aware strategy to enhance the matching and updating procedures for individual trajectories, and \citep{OAMOT} utilizes segmentation guidance to improve object-level ReID feature extraction.
Although these methods aim to improve the discriminative capability of visual representations, they overlook the fact that tracking only needs to distinguish objects within a finite set, rather than from a potentially unbounded identity space as in conventional ReID. 
This observation serves as the key motivation of our work, as it makes it possible to use tracking histories to guide the optimization of the feature space.
To the best of our knowledge, this is the first work to provide an intuitive explanation, a systematic analysis, and a practical solution to this issue.

\textbf{End-to-End MOT} models are emerging forces, bypassing hand-crafted algorithms \citep{ByteTrack, OC-SORT} to formulate multi-object tracking in an end-to-end manner \citep{MOTR, MOTIP}.
A typical form is to expand DETR \citep{DETR, DeformableDETR} into MOT tasks, representing different trajectories through the propagation of track queries \citep{MOTR, TrackFormer}.
Follow-up methods incorporated temporal information \citep{MeMOT, MeMOTR, SambaMOTR} and mitigated the imbalance of supervision signals \citep{MOTRv2, CO-MOT}, leading to better tracking performance.
Nevertheless, end-to-end methods still face the challenges of high computational costs and a strong need for training data, which will require future research.
Although beyond our main focus, this paradigm may also benefit from our findings, as discussed in the conclusion.
\section{Preliminary}
\label{sec:preliminary}

\subsection{ReID-based Tracker}
\label{sec:preliminary-tracker}

The tracking-by-detection paradigm \citep{SORT, ByteTrack, OC-SORT} treats multiple object tracking as a two-step process. First, an object detector $\detectModel$ is employed to localize all targets in a given frame $I_t$. Subsequently, these detections are associated with established trajectories based on a cost matrix or used to initialize new tracks.
Following our discussion in \cref{sec:introduction}, we simplify our experimental scope by concentrating on trackers that use only appearance cues for data association. Given an object bounding box, $\bboxWithti{t}{i}$, in the $t$-th frame, a feature extraction network $\reidModel$ is applied to obtain the corresponding visual feature $\featWithti{t}{i}$, often referred to as a re-identification (ReID) feature. 
It is used to represent the appearance of each detection and to construct the feature of each trajectory. In practice, while numerous methods \citep{Deep-SORT, Deep-OC-SORT, Hybrid-SORT} for trajectory modeling exist, we adopt the widely-used Exponential Moving Average (EMA) update strategy due to its proven efficiency and effectiveness, as formulated below:

\begin{equation}
    \featEMAWithtj{t}{j} = \lambda \featWithti{t}{\tau_j} + (1 - \lambda) \featEMAWithtj{t-1}{j},
    \label{eq:ema}
\end{equation}
where $\featEMAWithtj{t-1}{j}$ represents the appearance feature of track $\tau_j$ aggregated up to timestep $t-1$, $\featWithti{t}{\tau_j}$ is the ReID feature obtained from the extractor $\reidModel$ at the current frame $I_t$, and $\lambda$ is a momentum coefficient, typically set to a small value close to $0$, that controls the update ratio.

Once the aforementioned features are prepared, we compute the matching cost for each detection-trajectory pair using a similarity metric. A common practice is to use the cosine similarity, which is calculated as follows:

\begin{equation}
    \text{Cost}(t, i, \tau_j) = 1 - \text{Sim}(t, i, \tau_j) = 1 - \frac{\featWithti{t}{i} \cdot \featEMAWithtj{t-1}{j}}{\Vert \featWithti{t}{i} \Vert \Vert \featEMAWithtj{t-1}{j} \Vert}.
    \label{eq:cost}
\end{equation}
Accordingly, a cost matrix is constructed for the current frame based on all potential assignments. The Hungarian algorithm is then employed to find the globally optimal matching solution. 
Following this, the features of the matched tracks are updated according to \cref{eq:ema}, in preparation for the next time step.

\begin{figure}[tb] 
    \centering 
    \begin{minipage}{0.48\textwidth} 
        \centering 
        \includegraphics[width=0.95\linewidth]{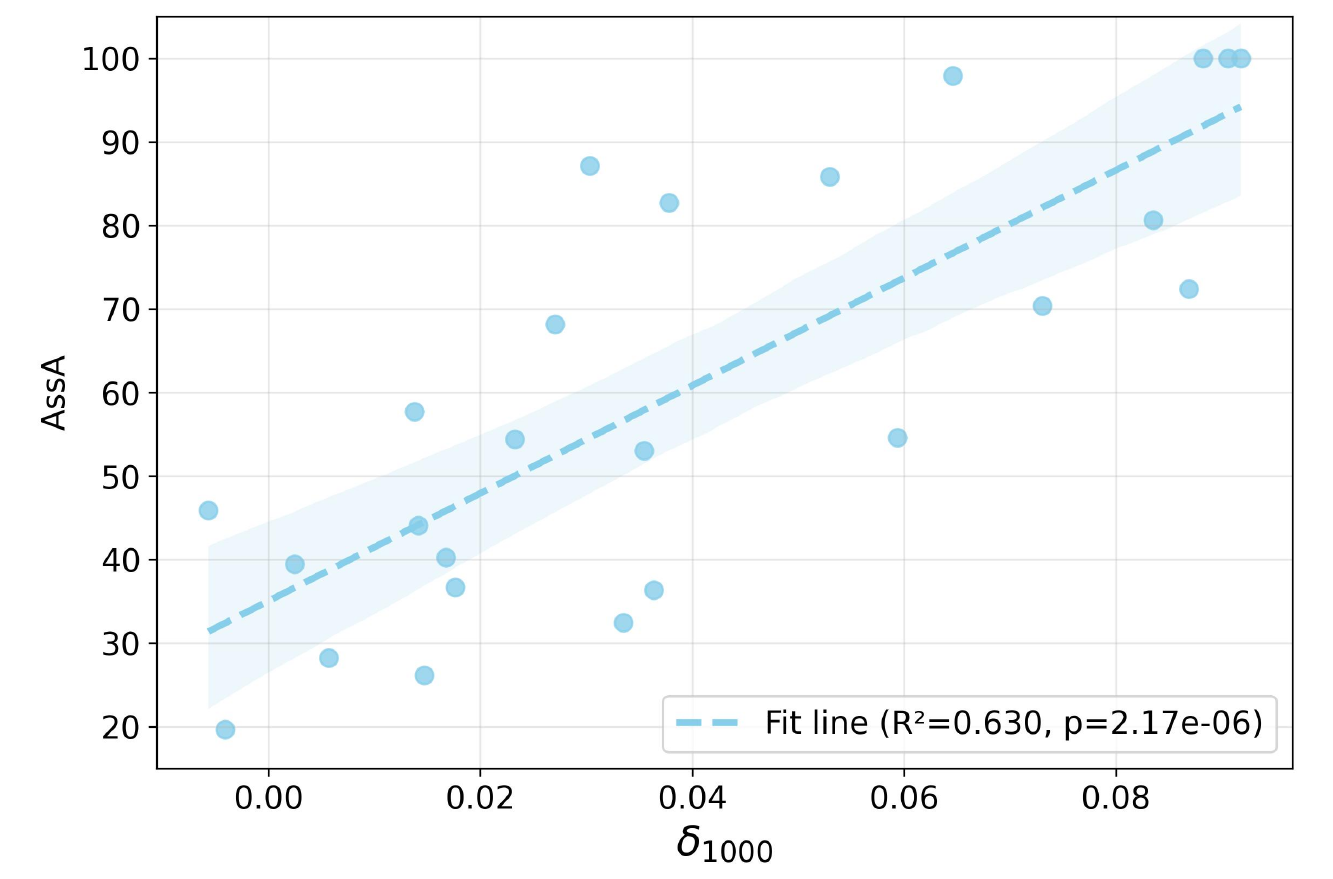} 
        \subcaption{Significant and reliable positive correlation between discriminability and tracking ability.}
        \label{fig:analysis-dancetrack}
    \end{minipage}
    \hfill 
    \begin{minipage}{0.48\textwidth} 
        \centering
        \includegraphics[width=0.95\linewidth]{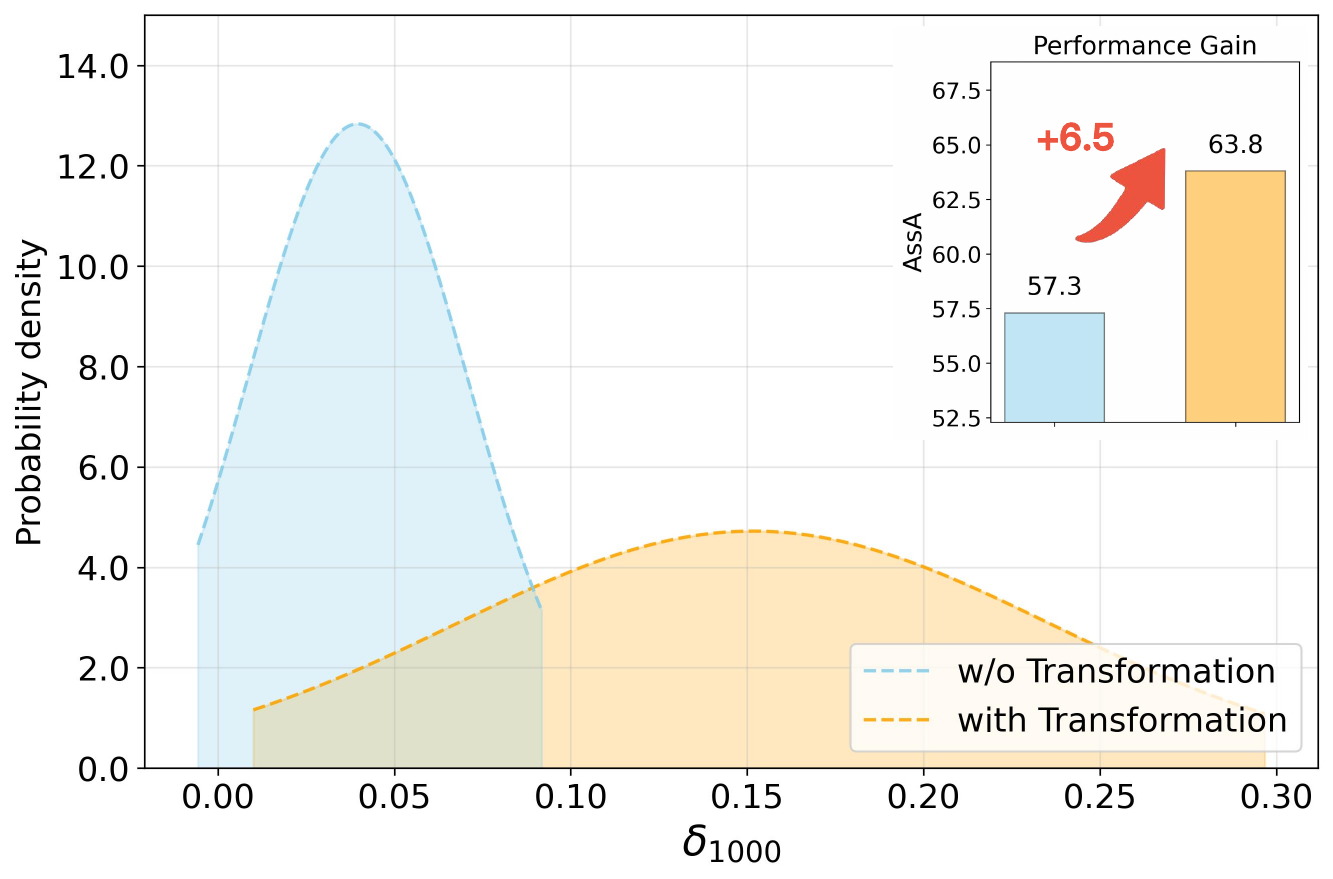}
        \subcaption{Our transformation improves performance by enhancing feature discriminative capability.}
        \label{fig:transformation-dancetrack}
    \end{minipage}
    \caption{Correlation between ReID feature discriminability $\delta_{1000}$ and tracking accuracy AssA on DanceTrack \citep{DanceTrack}.
    Similar analysis on \citet{SportsMOT} can be found in Appendix Fig.5.
    }
    \label{fig:analysis-and-transformation-dancetrack}
\end{figure}

\subsection{Discriminative Capability Analysis}
\label{sec:preliminary-analysis}

As stated in \cref{sec:preliminary-tracker}, since the tracker relies solely on appearance features for discrimination, it is intuitive to assume that the discriminative power of the ReID features is directly correlated with the tracking performance.

To validate this hypothesis, it is necessary to quantify the discriminative capability of the representation space. Since the tracking process relies on cosine similarity for affinity measurement, as shown in \cref{eq:cost}, we also adopt it as the cornerstone for evaluating the discriminability.
To be more specific, we measure the discriminative power for the $i$-th detection in frame $t$ using a score, $\delta(t, i)$. This score is defined as the similarity margin between the detection's positive track and its most confusing negative track. 
Furthermore, since tracking failures are minority events within a given sequence, we focus on the most challenging cases. Therefore, for each video, we select the $1000$ worst scores and compute their average. This metric, termed $\delta_{1000}$, is used to quantify the discriminative ability of the ReID representations for a specific sequence (details in Appendix A.1).
Accordingly, we conduct a statistical analysis on the representative dataset DanceTrack \citep{DanceTrack}, as shown in \cref{fig:analysis-dancetrack}. The results reveal a significant and reliable positive correlation between the discriminative capability ($\delta_{1000}$) of the ReID features and the object association accuracy (AssA \citep{HOTA}).
This conclusion provides a clear motivation for our work: to boost tracking performance by explicitly enhancing the discriminability of the representation space, described in \cref{sec:method}.

\subsection{Fisher Linear Discriminant}
\label{sec:preliminary-FLD}

Fisher Linear Discriminant (FLD) \citep{FLD}, also widely known as Linear Discriminant Analysis (LDA), is a classic supervised method used for both dimensionality reduction and classification. The core principle is to find a linear transformation that projects high-dimensional data onto a lower-dimensional space where the classes are maximally separated. In other words, the projection pulls the means of different classes far apart while keeping the data within each class tightly clustered.
Mathematically, this is achieved by defining a within-class scatter matrix, ${\displaystyle \mS}_W$, and a between-class scatter matrix, ${\displaystyle \mS}_B$. Given a set of $N$ feature vectors $\{ \displaystyle \vx_1, \displaystyle \vx_2, \cdots, \displaystyle \vx_N \} = \displaystyle \mX \in \displaystyle \sR^{N \times d}$, each feature $\displaystyle \vx$ is associated with one of $C$ classes, the scatter matrices can be formulated as:

\begin{equation}
    \displaystyle \mS_W = \sum_{c=1}^C \sum_{\displaystyle \vx \in \displaystyle \mX_c} (\displaystyle \vx - \bar{\displaystyle \vx}_c) {(\displaystyle \vx - \bar{\displaystyle \vx}_c)}^T ,
\label{eq:SW}    
\end{equation}

\begin{equation}
    \displaystyle \mS_B = \sum_{c=1}^C N_c (\bar{\displaystyle \vx}_c - \bar{\displaystyle \vx}) {(\bar{\displaystyle \vx}_c - \bar{\displaystyle \vx})}^T ,
\label{eq:SB}    
\end{equation}

\begin{equation}
    \bar{\displaystyle \vx} = \frac{1}{N} \sum_{i=1}^N \displaystyle \vx_i ,
    \qquad
    \bar{\displaystyle \vx}_c = \frac{1}{N_c} \sum_{\displaystyle \vx \in \displaystyle \mX_c} \displaystyle \vx ,
\label{eq:x-means}
\end{equation}
where $\displaystyle \mX_c$ represents the subset of $\displaystyle \mX$ pertaining to class $c$.
The optimal projection matrix, $\displaystyle \mW \in \displaystyle \sR^{d \times d'}$, is found by maximizing the Fisher criterion \citep{FLD}, which is the ratio of the between-class scatter to the within-class scatter in the projected space:

\begin{equation}
    J (\displaystyle \mW) = \frac{\displaystyle \mW^T \displaystyle \mS_B \displaystyle \mW}{\displaystyle \mW^T \displaystyle \mS_W \displaystyle \mW}.
\label{eq:fisher-criterion}
\end{equation}
By applying the projection matrix $\displaystyle \mW$ derived above, each feature $\displaystyle \vx$ is converted into a new $d'$-dimensional vector with enhanced discriminability, where $d' = \min(C-1, d)$.

\section{Method}
\label{sec:method}

\begin{figure}[tb]
    \centering
    \includegraphics[width=0.95\linewidth]{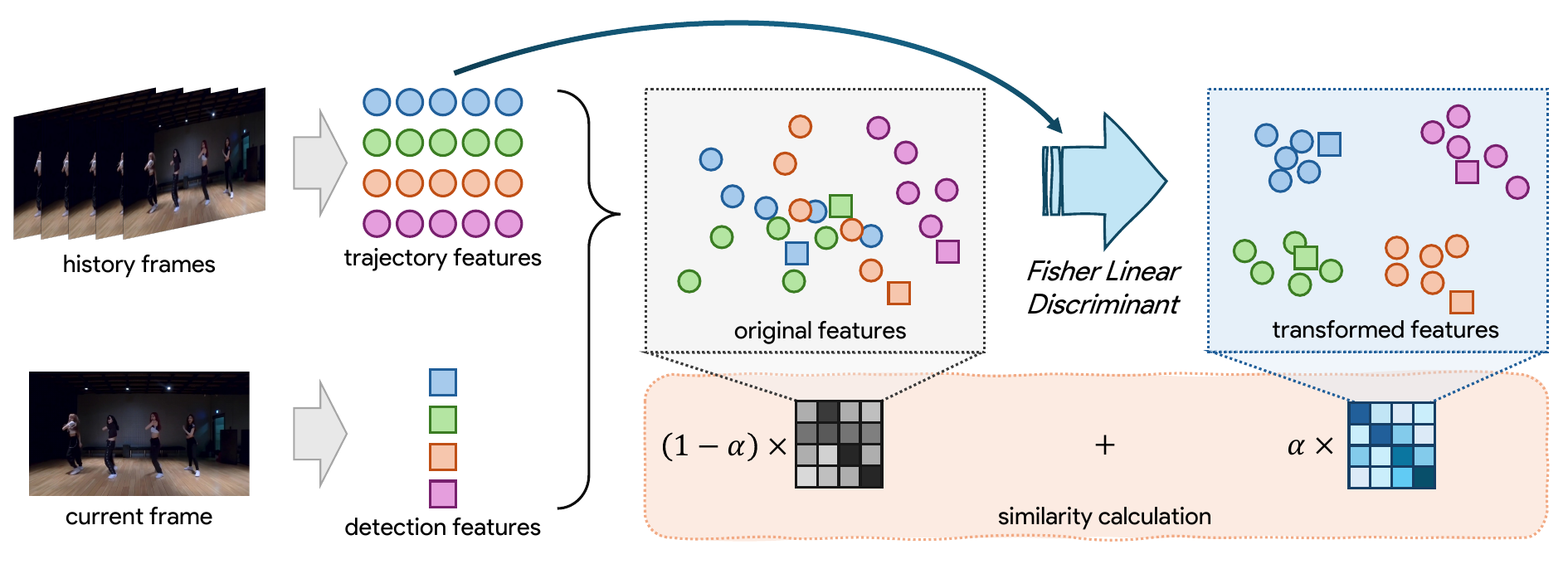}
    \caption{
    \textbf{Overview of our pipeline.}
    We use different colors to indicate different identities (trajectories).
    In the original space, some overly similar targets cannot be well distinguished, leading to issues in the matching process.
    Therefore, we treat the trajectory features as conditions and apply a tailored \textit{Fisher Linear Discriminant} to seek a better subspace for distinguishing different trajectories. 
    Finally, both original and transformed features are used to calculate the similarity matrix, balancing generalization and specialization.
    }
    \label{fig:overview}
\end{figure}

Based on the analysis in \cref{sec:preliminary-analysis} and the result shown in \cref{fig:analysis-dancetrack}, a clear positive correlation exists between the discriminative capability of the ReID features and the final tracking performance. 
Therefore, in this section, our primary goal is to find a more discriminative representation space for distinguishing between trajectories. 
To this end, we mainly employ Fisher Linear Discriminant (FLD) \citep{FLD} along with several customized techniques, which are detailed in \cref{sec:method-transformation} and \cref{sec:method-temporal} - \cref{sec:method-integration}, respectively. The overall illustration is shown in \cref{fig:overview}.

\subsection{History-Aware Transformation for ReID Features}
\label{sec:method-transformation}

As discussed in \cref{sec:introduction}, current multi-object tracking (MOT) methods \citep{Deep-OC-SORT, Hybrid-SORT, DiffMOT} largely adopt ReID features directly from traditional re-identification methods \citep{ReID, FastReID}. 
Since these models are required to distinguish between a vast number of open-set identities, the features they produce are, by design, as general as possible.
In contrast, the multiple object tracking task only requires recognizing a closed set of identities within a single video. This creates a dilemma where the generality of traditional ReID features becomes a liability, as they lack the specificity needed to differentiate between these similar targets, as illustrated in \cref{fig:sequence}.
Therefore, we are motivated to seek a more specialized representation space to address the aforementioned challenges.
Intuitively, this space should pull features belonging to the same trajectory closer, while pushing features from different trajectories further apart.
This idea coincides perfectly with the objective of Fisher Linear Discriminant (FLD) \citep{FLD} in its mathematical formulation, provided that we treat each \textit{trajectory} as a \textit{class} in the original framework. Specifically, by replacing the feature vector $\displaystyle \vx$ in \cref{eq:SW} - \cref{eq:x-means} with our ReID features $\displaystyle \vf$, and substituting the number of classes $C$ with the number of tracks $N_\tau$, we can obtain the projection matrix $\displaystyle \mW$ for tracking by maximizing the objective in \cref{eq:fisher-criterion}.

However, FLD is a supervised method, which means it requires corresponding labels in addition to the feature vectors. This core prerequisite is unfulfilled in a standard tracking process.
Therefore, we propose a history-aware dynamic labeling scheme to compensate for this absence.
Practically, since tracking is an online process, at each timestep $t$, the historical track assignments from previous frames can serve as the supervisory signals for FLD. Although potential tracking errors exist, we believe the overall statistical signal remains reliable.
Furthermore, since a target's appearance gradually evolves during tracking, we only consider its $T$ most recent features for each trajectory. This choice ensures both efficiency and effectiveness.

\subsection{Temporally-Weighted Trajectory Centroid}
\label{sec:method-temporal}

Following the statement in \cref{sec:method-transformation}, a naive implementation would be to average all $T$ features $\{ \featWithti{t-T}{\tau_j}, \cdots, \featWithti{t-2}{\tau_j}, \featWithti{t-1}{\tau_j} \}$ of the $\tau_j$-th trajectory to serve as its mean feature center. According to the definition of FLD \citep{FLD} and \cref{eq:SB}, these feature centroids determine the distribution centers of the vectors after projection.
Although this approach yields notable improvements, we still point out that it overlooks the temporal characteristics inherent in the tracking task.
In online tracking, a target's appearance evolves continuously over time. Even within the same trajectory, features that are closer temporally tend to have higher similarity. Hence, for identity allocation at the current moment, more recent ReID features should intuitively play a more significant role. 
In practice, we apply a temporal weighting to the mean calculation in \cref{eq:x-means}:

\begin{equation}
    \bar{\feat} = \frac{1}{N_\tau} \sum_{j=1}^{N_\tau} \bar{\feat}_{\tau_j},
    \qquad
    \bar{\feat}_{\tau_j} = \frac{1}{\sum \lambda_{t'}} \sum_{t'=t-T}^{t-1} \lambda_{t'} \featWithti{t'}{\tau_j},
    \qquad
    \lambda_{t'} = (\lambda_0)^{t-t'},
\label{eq:weighted-means}
\end{equation}
where $\lambda_0$ is a temporal decay coefficient with a value between $0$ and $1$. 
Using these temporal-weighted trajectory centroids in the calculation of \cref{eq:SB} makes the final projection more attuned to the current temporal context, benefiting the similarity measurement at the time step $t$.

\subsection{Knowledge Integration}
\label{sec:method-integration}

Although we have found a more discriminative space conditioned on historical trajectories with the methods in \cref{sec:method-transformation} and \cref{sec:method-temporal}, it still has some imperfections.
First, the historical tracking results may contain errors, which can lead to a biased or suboptimal projection matrix. Second, because the transformed space is built only from the features of existing trajectories, it may not be robust enough for handling newborn targets.
Therefore, we revisit the original representation space. Although it is not optimized for a given scenario, it offers more robust generalization capabilities, especially when facing unseen targets. This motivates our proposal to integrate it with the specialized subspace for a trade-off.
Due to the disparate dimensionalities of these two spaces, our integration strategy operates on the similarity matrices rather than the vectors themselves. It can be formulated as follows:

\begin{equation}
    \text{Cost}^*(t, i, \tau_j) = 1 - \text{Sim}^*(t, i, \tau_j) = 1 - [\alpha \cdot \text{Sim}'(t, i, \tau_j) + (1-\alpha) \cdot \text{Sim}(t, i, \tau_j)],
    \label{eq:integrate-cost}
\end{equation}
where $\text{Sim}'(\cdot)$ is the similarity computed using the transformed ReID features, and $\alpha$ is a balancing coefficient. 
The Hungarian algorithm then finds the optimal assignment using the complete cost matrix constructed from the fused $\text{Cost}^*(\cdot)$.
See \cref{fig:overview} for an overview of this pipeline.

\section{Experiments}
\label{sec:experiments}

\subsection{Datasets and Metrics}
\label{sec:experiments-datasets-and-metrics}

\textbf{Datasets.} 
We select DanceTrack \citep{DanceTrack} and SportsMOT \citep{SportsMOT} as our primary experimental benchmarks because they both present a key challenge: targets within a single video often exhibit a high degree of visual similarity. Specifically, DanceTrack features group dance scenarios, while SportsMOT includes three types of team sports.
We also evaluate our approach on the TAO \citep{TAO} dataset to demonstrate its effectiveness in diverse and general tracking cases.
In addition, we report compact MOT17 \cite{MOT16} results in \ref{tab:sota-mot17} and detailed results in Appendix B.1.

\noindent \textbf{Metrics.}
On traditional MOT benchmarks \citep{MOT16, DanceTrack, SportsMOT}, we select the Higher Order Tracking Accuracy (HOTA) \citep{HOTA} as the primary metric, especially its Association Accuracy (AssA) component. We also include MOTA \citep{MOTA} and IDF1 \citep{IDF1} in some experiments.
To better evaluate the multi-category tracking problem, we employ the Tracking Every Thing Accuracy (TETA) \citep{TETer} for TAO \citep{TAO}.

\subsection{Implementation Details}
\label{sec:experiments-implementation-details}

To more clearly illustrate the improvements brought by our method, we focus our experiments on pure ReID-based trackers, as discussed in \cref{sec:preliminary-tracker}. Due to the lack of such publicly available trackers in the community, we construct a new tracker by combining the widely-used YOLOX \citep{YOLOX} detector with the FastReID \citep{FastReID} model. 
For a fair comparison, we use the off-the-shelf weights from \citet{OC-SORT, Hybrid-SORT, DiffMOT} for all network modules without any additional fine-tuning. To ensure the baseline performs at its full potential, we conduct an exhaustive grid search to identify the optimal hyperparameter configuration for each specific benchmark.
This fortified baseline, denoted as \textit{FastReID-MOT}, ensures that any subsequent performance gains are strictly attributable to our proposed methodology.
To further evaluate the potential of our approach, we select several representative hybrid-based trackers \cite{ByteTrack, OC-SORT, Hybrid-SORT} and incorporate our method into their ReID branches to boost performance.
Regarding MASA \cite{MASA}, we also directly adopt the official model weights without any modifications, covering a diverse range of backbones to guarantee the breadth of our evaluation.
For notation, we add the prefix \textbf{HAT-} to methods that use our \textbf{H}istory-\textbf{A}ware \textbf{T}ransformation approach.

\begin{table}[tb]
  \begin{minipage}[t]{.5\textwidth}
    \centering
\caption{
Comparison with state-of-the-art methods on the DanceTrack test set.
}
\label{tab:sota-dancetrack}
\footnotesize
\setlength{\tabcolsep}{1.5pt}{
    \begin{tabular}{l|ccccc}
      \toprule[2pt]
      Methods & HOTA & DetA & AssA \\
      \midrule[1pt]
      \textit{\scriptsize \underline{motion-based:}} \\
      \scriptsize ByteTrack \tiny \citep{ByteTrack} & 47.7 & 71.0 & 32.1 \\
      \scriptsize OC-SORT \tiny \citep{OC-SORT} & 55.1 & 80.3 & 38.3 \\
      \scriptsize C-BIoU \tiny \citep{C-BIoU} & 60.6 & 81.3 & 45.4 \\
      \midrule[0.5pt]
      \textit{\scriptsize \underline{ReID-based:}} \\
      \scriptsize QDTrack \tiny \citep{QDTrack} & 54.2 & 80.1 & 36.8 \\
      \scriptsize FastReID-MOT \tiny (baseline) & 50.6 & 81.1 & 31.6 \\
      \scriptsize HAT-FastReID-MOT & 58.6 & 81.3 & 42.3 \\
      \scriptsize HAT-FastReID-MOT$\dagger$ & \bf 61.2 & \bf 81.6 & \bf 46.0 \\
      \midrule[0.5pt]
      \textit{\scriptsize \underline{hybrid-based:}} \\
      \scriptsize FairMOT \tiny \citep{FairMOT} & 39.7 & 66.7 & 23.8 \\
      \scriptsize DeepSORT \tiny \citep{Deep-SORT} & 45.6 & 71.0 & 29.7 \\
      \scriptsize StrongSORT \tiny \citep{StrongSORT} & 55.6 & 80.7 & 38.6 \\
      \scriptsize DiffMOT \tiny \citep{DiffMOT} & 62.3 & \bf 82.5 & 47.2 \\
      \scriptsize Hybrid-SORT-ReID \tiny \citep{Hybrid-SORT} & 65.7 & -- & -- \\
      \scriptsize ByteTrack-ReID & 52.4 & 71.0 & 38.7 \\
      \scriptsize HAT-ByteTrack-ReID & 56.1 & 71.4 & 44.2 \\
      \scriptsize OC-SORT-ReID & 60.8 & 81.0 & 45.7 \\
      \scriptsize HAT-OC-SORT-ReID & 64.6 & 81.5 & 51.3 \\
      \scriptsize HAT-Hybrid-SORT-ReID & \bf 66.9 & 81.5 & \bf 55.0 \\
      \midrule[0.5pt]
      \textit{\scriptsize \textcolor{gray}{\underline{end-to-end:}}} \\
      \scriptsize \textcolor{gray}{MeMOTR} \tiny \citep{MeMOTR} & \textcolor{gray}{63.4} & \textcolor{gray}{77.0} & \textcolor{gray}{52.3} \\
      \scriptsize \textcolor{gray}{CO-MOT} \tiny \citep{CO-MOT} & \textcolor{gray}{65.3} & \textcolor{gray}{80.1} & \textcolor{gray}{53.5} \\
      \scriptsize \textcolor{gray}{SambaMOTR} \tiny \citep{SambaMOTR} & \textcolor{gray}{67.2} & \textcolor{gray}{78.8} & \textcolor{gray}{57.5} \\
      \scriptsize \textcolor{gray}{MOTIP} \tiny \citep{MOTIP} & \bf \textcolor{gray}{69.6} & \bf \textcolor{gray}{80.4} & \bf \textcolor{gray}{60.4} \\
      \bottomrule[2pt]
    \end{tabular}
  }

  \end{minipage}%
  \hfill
  \begin{minipage}[t]{.48\textwidth}
    \centering
\caption{
Performance on the SportsMOT test set.
Results marked with * denote joint training involving the validation set of SportsMOT.
}
\label{tab:sota-sportsmot}
\footnotesize
\setlength{\tabcolsep}{1.5pt}{
    \begin{tabular}{l|ccccc}
      \toprule[2pt]
      Methods & HOTA & DetA & AssA \\
      \midrule[1pt]
      \textit{\scriptsize \underline{motion-based:}} \\
      \scriptsize ByteTrack \tiny \citep{ByteTrack} & 62.8 & 77.1 & 51.2 \\
      \scriptsize OC-SORT \tiny \citep{OC-SORT} & 71.9 & 86.4 & 59.8 \\
      \scriptsize \textcolor{gray}{ByteTrack* \tiny \citep{ByteTrack}} & \textcolor{gray}{64.1} & \textcolor{gray}{78.5} & \textcolor{gray}{52.3} \\
      \scriptsize \textcolor{gray}{OC-SORT* \tiny \citep{OC-SORT}} & \textcolor{gray}{73.7} & \textcolor{gray}{88.5} & \textcolor{gray}{61.5} \\
      \midrule[0.5pt]
      \scriptsize \textit{\underline{ReID-based:}} \\
      \scriptsize QDTrack \citep{QDTrack} & 60.4 & 77.5& 47.2 \\
      \scriptsize FastReID-MOT \tiny (baseline) & 67.3 & 86.8 & 52.3 \\
      \scriptsize HAT-FastReID-MOT & 78.1 & 87.3 & 69.9 \\
      \scriptsize HAT-FastReID-MOT$\dagger$ & \bf 78.9 & \bf 87.4 & \bf 71.3 \\
      \scriptsize \textcolor{gray}{HAT-FastReID-MOT$\dagger$*} & \textcolor{gray}{80.8} & \textcolor{gray}{89.4} & \textcolor{gray}{73.1} \\
      \midrule[0.5pt]
      \textit{\scriptsize \underline{hybrid-based:}} \\
      \scriptsize BoT-SORT \tiny \citep{BoT-SORT} & 68.7 & 84.4 & 55.9 \\
      \scriptsize DiffMOT \tiny \citep{DiffMOT} & 72.1 & 86.0 & 60.5 \\
      \scriptsize ByteTrack-ReID & 65.1 & 76.8 & 55.1 \\
      \scriptsize HAT-ByteTrack-ReID & 72.4 & 77.3 & 67.8 \\
      \scriptsize OC-SORT-ReID & 74.1 & 86.8 & 63.3 \\
      \scriptsize HAT-OC-SORT-ReID & \bf 81.2 & \bf 87.2 & \bf 75.6 \\
      \scriptsize \textcolor{gray}{HAT-OC-SORT-ReID*} & \textcolor{gray}{82.4} & \textcolor{gray}{89.3} & \textcolor{gray}{76.1} \\
      
      \midrule[0.5pt]
      \textit{\scriptsize \textcolor{gray}{\underline{end-to-end:}}} \\
      \scriptsize \textcolor{gray}{TrackFormer} \tiny \citep{TrackFormer} & \textcolor{gray}{63.3} & \textcolor{gray}{66.0} & \textcolor{gray}{61.1} \\
      \scriptsize \textcolor{gray}{MeMOTR} \tiny \citep{MeMOTR} & \textcolor{gray}{68.8} & \textcolor{gray}{82.0} & \textcolor{gray}{57.8} \\
      \scriptsize \textcolor{gray}{MOTIP} \tiny \citep{MOTIP} & \bf \textcolor{gray}{72.6} & \bf \textcolor{gray}{83.5} & \bf \textcolor{gray}{63.2} \\
      \bottomrule[2pt]
    \end{tabular}
  }

  \end{minipage}
\end{table}

\subsection{State-of-the-art Comparison}
\label{sec:experiments-sota}

\noindent \textbf{FastReID-MOT.}
We compare our method (HAT-FastReID-MOT) against the baseline (FastReID-MOT) on DanceTrack \cite{DanceTrack} and SportsMOT \cite{SportsMOT} in \cref{tab:sota-dancetrack} and \cref{tab:sota-sportsmot}.
$\dagger$ indicates that hyperparameters are fine-tuned on the corresponding dataset to maximize performance; otherwise, the default settings from our ablation study are used, as stated in \cref{sec:experiments-ablation}.
Our approach yields substantial performance gains over the baseline.
On the challenging DanceTrack dataset, our appearance-only method even achieves results comparable to several recent hybrid and motion-based trackers \citep{C-BIoU, DiffMOT}.
Even more impressively, on the SportsMOT benchmark, our ReID-only tracker (HAT-FastReID-MOT) sets a new state-of-the-art. It significantly surpasses existing hybrid-based trackers \cite{DiffMOT} and end-to-end models \cite{SambaMOTR, MOTIP}.
This result both vindicates our approach and highlights the need to reconsider the true potential of ReID features.


\noindent \textbf{Hybrid-based Tracker.}
To further validate the effectiveness of our method, we inserted it into several recent well-known trackers \citep{ByteTrack, OC-SORT, Hybrid-SORT}. 
The results in \cref{tab:sota-dancetrack} and \cref{tab:sota-sportsmot} show that our method can consistently bring significant improvements when applied to hybrid-based trackers.
While many prior studies \cite{MOTR, MeMOTR, SambaMOTR} suggest that end-to-end methods possess an inherent advantage on DanceTrack due to their flexibility, our approach surpasses several recent E2E models \cite{CO-MOT, SambaMOTR}, demonstrating significant potential. A more detailed discussion regarding E2E methods is provided in Appendix C.
The smaller performance gains on hybrid-based methods can be attributed to both performance saturation and the inherent design of these trackers, which often prioritizes motion and thus limits the impact of our appearance enhancements.
In addition, the complex design of collaborative trackers also limits the improvement brought by a single module and makes it challenging to coordinate different modules. 
To further exploit the capability of the proposed approach in hybrid-based methods, we believe that it may be necessary to redesign the fusion algorithm of the tracker itself and conduct extensive engineering-level parameter tuning, which is beyond the scope of this work.

\begin{table}[tb]
  \centering
  \caption{
  Evaluating our method by applying it to MASA \citep{MASA}.
  All models are trained on a large-scale image segmentation dataset \citep{SAM} with different visual backbones.
  }
  \label{tab:sota-masa}
  \footnotesize
  \setlength{\tabcolsep}{3pt}{
    \begin{tabular}{l|cc|cc|cc}
      \toprule[2pt]
       \multirow{2}{*}{Methods} & \multicolumn{2}{c|}{DanceTrack test} & \multicolumn{2}{c|}{SportsMOT test} & \multicolumn{2}{c}{TAO val}\\
       & HOTA & AssA & HOTA & AssA & TETA & AssocA \\
      \midrule[1pt]
      \multicolumn{1}{l|}{\scriptsize \textit{\underline{MASA \citep{MASA}:}}} \\
       MASA-R50 & 50.8 & 31.6 & 71.6 & 58.9 & 45.8 & 42.7 \\
       MASA-Detic & 50.6 & 31.5 & 72.2 & 60.1 & 46.5 & 44.5 \\
       MASA-G-DINO & 50.4 & 31.2 & 72.8 & 61.0 & 46.8 & 45.0 \\
       MASA-SAM-B & 49.4 & 29.9 & 71.9 & 59.5 & 46.2 & 43.7 \\
      \midrule[0.5pt]
      \multicolumn{1}{l|}{\scriptsize \textit{\underline{Ours (HAT):}}} \\
       MASA-R50 & 54.3 \improve{+3.5} & 36.1 \improve{+4.5} & 73.7 \improve{+2.1} & 62.4 \improve{+3.5} & 46.4 \improve{+0.6} & 44.4 \improve{+1.7}\\
       MASA-Detic & 54.3 \improve{+3.7} & 36.2 \improve{+5.7} & 74.5 \improve{+2.3} & 63.7 \improve{+3.6} & 47.2 \improve{+0.7} & 46.4 \improve{+1.9} \\
       MASA-G-DINO & 53.9 \improve{+3.5} & 35.7 \improve{+4.5} & 74.7 \improve{+1.9} & 64.1 \improve{+3.1} & 47.5 \improve{+0.7} & 46.7 \improve{+1.7} \\
       MASA-SAM-B & 52.1 \improve{+2.7} & 33.4 \improve{+3.5} & 73.4 \improve{+1.5} & 61.9 \improve{+2.4} & 46.9 \improve{+0.7} & 45.6 \improve{+1.9} \\
      \bottomrule[2pt]
    \end{tabular}
  }
\end{table}

\begin{figure*}[tb]
  \centering
  \begin{minipage}[c]{0.50\linewidth}
    \centering
    \captionof{table}{
      Performance comparison with state-of-the-art methods on MOT17 \citep{MOT16}.
      The best and second-best results are denoted in \textbf{bold} and \underline{underline}, respectively.
    }
    \vspace{8pt}
    \label{tab:sota-mot17}
    \setlength{\tabcolsep}{3pt}
    \resizebox{\linewidth}{!}{%
    \begin{tabular}{l|cccc}
      \toprule[2pt]
      Methods & HOTA & DetA & AssA & IDF1 \\
      \midrule[1pt]
      \textit{\underline{reid-based:}} \\
      QDTrack \citep{QDTrack}                    & 53.9 & 55.6 & 52.7 & 66.3 \\
      FastReID-MOT (baseline)                    & \underline{61.5} & \underline{63.4} & \underline{60.0} & \underline{73.5} \\
      HAT-FastReID-MOT$\dagger$                  & \bf 63.5 & \bf 64.0 & \bf 63.2 & \bf 77.5 \\
      \midrule[0.5pt]
      \textit{\underline{hybrid-based:}} \\
      FairMOT \citep{FairMOT}                    & 59.3 & 60.9 & 58.0 & 72.3 \\
      DeepSORT \citep{Deep-SORT}                 & 61.2 & 63.1 & 59.7 & 74.5 \\
      DiffMOT \citep{DiffMOT}                    & \bf 64.5 & \bf 64.7 & \bf 64.6 & \bf 79.3 \\
      OC-SORT-ReID                               & 64.1 & \underline{64.4} & 64.0 & 79.0 \\
      HAT-OC-SORT-ReID                           & \underline{64.2} & \underline{64.4} & \underline{64.1} & \underline{79.2} \\
      \bottomrule[2pt]
    \end{tabular}%
    }
  \end{minipage}
  \hfill
  \begin{minipage}[c]{0.46\linewidth}
    \centering
    \includegraphics[width=\linewidth]{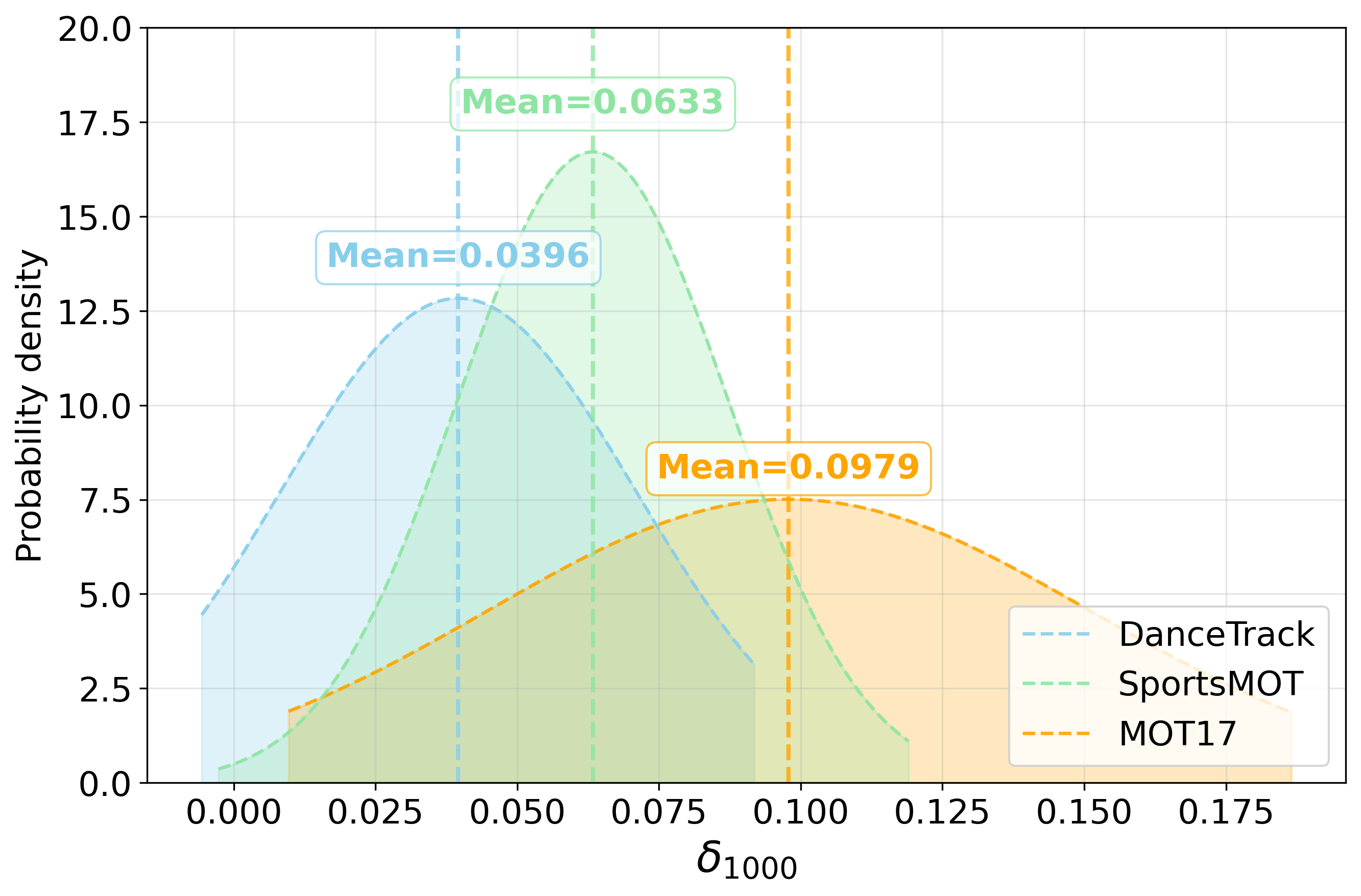}
    \captionof{figure}{
      Comparison of ReID separability in the original feature space on DanceTrack \citep{DanceTrack},
      SportsMOT \citep{SportsMOT}, and MOT17 \citep{MOT16} based on $\delta_{1000}$.
    }
    \label{fig:mot17-distribution}
  \end{minipage}
\end{figure*}

\begin{table}[tb]
  \centering
  \caption{
  Comparison of transformation selections. 
  \textit{Oracle} and \textit{YOLOX} denote the sources of the detection results, while $d$ and $d'$ indicate the original and projected feature dimension, respectively.
  $N_\textit{obj}$ and $N_\textit{id}$ are the total number of historical samples and trajectories, respectively.
  If $d' > d$, the target dimension will be set to $d$.
  }
  \label{tab:transformation}
  \footnotesize
  \setlength{\tabcolsep}{3pt}{
    \begin{tabular}{c|c|c|c|cc|cc}
      \toprule[2pt]
      \multirow{2}{*}{\#} & \multirow{2}{*}{$\detectModel$} & \multirow{2}{*}{\textit{Method}} & \multirow{2}{*}{$d'$} & \multicolumn{2}{c|}{DanceTrack val} & \multicolumn{2}{c}{SportsMOT val} \\
       & & & & HOTA & AssA & HOTA & AssA \\
      \midrule[1pt]
       \# 1 & \multirow{4}{*}{\textit{Oracle}} & -- & $d$ & 74.9 & 57.3 & 86.2 & 74.7 \\
       \# 2 & & \textit{PCA} & $N_{\textit{obj}} - 1$ & 75.1 \improve{+0.2} & 57.6 \improve{+0.3} & 85.6 \drop{-0.6} & 73.9 \drop{-0.8} \\
       \# 3 & & \textit{PCA} & $N_{\textit{id}} - 1$ & 56.3 \drop{-18.6} & 32.5 \drop{-24.8} & 69.4 \drop{-16.8} & 49.0 \drop{-25.7} \\
       \# 4 & & \textit{FLD} & $N_{\textit{id}} - 1$ & \bf 79.0 \improve{+4.1} & \bf 63.8 \improve{+6.5} & \bf 92.2 \improve{+6.0} & \bf 85.4 \improve{+10.7} \\
       \midrule[1pt]
       \# 5 & \multirow{4}{*}{\textit{YOLOX}} & -- & $d$ & 51.1 & 33.4 & 73.7 & 61.5 \\
       \# 6 & & \textit{PCA} & $N_{\textit{obj}} - 1$ & 45.3 \drop{-5.8} & 26.7 \drop{-6.7} & 70.6 \drop{-3.1} & 56.4 \drop{-5.1} \\
       \# 7 & & \textit{PCA} & $N_{\textit{id}} - 1$ & 43.5 \drop{-7.6} & 24.6 \drop{-8.8} & 61.3 \drop{-12.4} & 42.7 \drop{-18.8} \\
       \# 8 & & \textit{FLD} & $N_{\textit{id}} - 1$ & \bf 57.7 \improve{+6.6} & \bf 42.7 \improve{+9.3} & \bf 81.1 \improve{+7.4} & \bf 74.1 \improve{+12.6} \\
      \bottomrule[2pt]
    \end{tabular}
  }
\end{table}

\noindent \textbf{MASA.}
To investigate the generalization of our method across different ReID representation spaces, we conduct experiments on MASA \citep{MASA}. This framework is ideal as it includes a variety of visual backbones \citep{ResNet, Detic, Grounding-DINO, SAM} and is pre-trained on a general-purpose segmentation dataset \citep{SAM}. 
To complement our evaluations on specialized datasets \cite{DanceTrack, SportsMOT}, we further employ the TAO \cite{TAO} benchmark. With its vast category diversity, TAO serves as a rigorous testbed to assess whether our approach maintains consistent reliability in open-world environments.
\cref{tab:sota-masa} shows that our approach brings consistent and significant boosts across all tested visual backbones.
However, the performance gains are less pronounced compared to those observed in \cref{tab:sota-dancetrack} and \cref{tab:sota-sportsmot}. We attribute this to the fact that our method is designed to refine and distill an existing feature space. Since MASA is not explicitly trained on specific datasets, its initial representations may possess limited discriminative power in challenging scenarios, providing a sub-optimal foundation for our distillation process.
Furthermore, the TAO \cite{TAO} dataset encompasses a vast array of heterogeneous object categories with minimal inter-class similarity, which naturally constrains the headroom for our algorithm. Nevertheless, the consistent improvements achieved across such diverse scenarios underscore the robust generalization of our method.

\noindent \textbf{MOT17.}
We provide a compact report of tracking performance on MOT17 \cite{MOT16} in \cref{tab:sota-mot17}.
Unlike the strong performance observed on DanceTrack and SportsMOT, our method achieves only moderate results on MOT17. Even on our self-constructed FastReID-MOT baseline, it brings an improvement of only $2.0$ HOTA.
As shown in \cref{fig:mot17-distribution}, we visualize the difference in discriminability within the original representation space across different datasets. MOT17 exhibits the highest appearance-level discriminability, likely due to the substantially different clothing of pedestrians. This property substantially limits the room for improvement of our proposed method.
This may also explain the marginal improvement on TAO \cite{TAO} in \cref{tab:sota-masa}, which is a practical consideration in real-world scenarios.

\subsection{Ablation Study}
\label{sec:experiments-ablation}

We verify the effectiveness of each component in this section, using the ReID-based tracker from \cref{sec:preliminary-tracker} and \cref{sec:experiments-implementation-details} as the baseline. For all experiments, except those in \cref{tab:transformation}, we use the detections from the public YOLOX model \citep{YOLOX, OC-SORT}.
We perform a step-by-step inquiry in \cref{tab:transformation} to \cref{tab:alpha}, with details provided in Appendix A.5. The \colorbox{gray!30}{gray cells} denote the final default hyperparameter settings.

\noindent \textbf{History-Aware Transformation.}
As shown in \cref{tab:transformation}, applying the FLD-based ReID feature transformation, as described in \cref{sec:method-transformation}, significantly improves tracking performance over the baseline (\# 1 and \# 5).
Following the correlation analysis in \cref{sec:preliminary-analysis}, we visualize the change in the discriminative ability $\delta_{1000}$ of ReID features under an oracle detection setting, as shown in \cref{fig:transformation-dancetrack}.
This serves as clear evidence that our history-aware transformation boosts the separability of visual representations, thereby improving tracking capabilities.
For comparison, we evaluate a PCA-based transformation in \cref{tab:transformation}, but it resulted in a performance drop. This is because Principal Component Analysis (PCA) is designed to maximize global variance and is oblivious to trajectory labels, which we believe are essential for finding an optimal space for tracking.

\begin{table}[tb]
  \label{tab:main_table_minipage}
  \begin{minipage}[t]{.45\textwidth}
    \centering
    \caption{
    History Length $T$
    }
    \label{tab:temporal-length}
    \small
    \setlength{\tabcolsep}{5pt}
    \begin{tabular}{c|cccc}
        \toprule[2pt]
        $T$ & HOTA & AssA & MOTA & IDF1 \\ 
        \midrule[1pt]
         $10$ & 54.3 & 37.8 & \bf 86.7 & 53.2 \\
         $20$ & 56.2 & 40.4 & 86.5 & 55.2 \\
         $40$ & 57.0 & 41.4 & 86.5 & 56.6 \\
         \rowcolor{gray!20} $60$ & \bf 57.7 & \bf 42.7 & 86.3 & \bf 56.7 \\
         $80$ & 56.3 & 40.5 & 86.2 & 55.4 \\
         $\infty$ & 55.3 & 39.2 & 85.1 & 52.2 \\
        \bottomrule[2pt]
    \end{tabular}
  \end{minipage}
  \hfill
  \begin{minipage}[t]{.52\textwidth}
    \centering
    \caption{
    Temporal Decay Coefficient $\lambda_0$
    }
    \label{tab:temporal-weight}
    \small
    \setlength{\tabcolsep}{5pt}
    \begin{tabular}{c|cccc}
        \toprule[2pt]
        $\lambda_0$ & HOTA & AssA & MOTA & IDF1 \\ 
        \midrule[1pt]
         1.00 & 57.7 & 42.7 & 86.3 & 56.7 \\
         0.95 & 58.5 & 42.5 & 86.8 & 58.2 \\
         \rowcolor{gray!20} 0.90 & \bf 59.3 & \bf 44.8 & 86.9 & \bf 59.8 \\
         0.80 & 59.1 & 44.7 & \bf 87.0 & 59.6 \\
         0.60 & 58.1 & 43.2 & \bf 87.0 & 57.9 \\
         0.40 & 57.8 & 42.8 & 86.9 & 57.1 \\
        \bottomrule[2pt]
    \end{tabular}
  \end{minipage}
\end{table}

\begin{table}[tb]
  \label{tab:more_ablation}
  \begin{minipage}[t]{.40\textwidth}
    \centering
    \caption{Balancing Coefficient $\alpha$.}
    \label{tab:alpha}
    \small
    \setlength{\tabcolsep}{2.5pt}
    \begin{tabular}{c|cccc}
    \toprule[2pt]
    $\alpha$ & HOTA & AssA & MOTA & IDF1 \\ 
    \midrule[1pt]
    1.0 & 59.3 & 44.8 & 86.9 & 59.8 \\
    \rowcolor{gray!20} 0.9 & \bf 60.6 & \bf 46.8 & 87.1 & \bf 61.7 \\
    0.8 & 59.3 & 45.0 & 87.2 & 60.4 \\
    0.6 & 59.1 & 44.7 & 87.6 & 60.6 \\
    0.4 & 58.7 & 43.9 & \bf 87.9 & 60.1 \\
    \bottomrule[2pt]
    \end{tabular}
  \end{minipage}
  \hfill
  \begin{minipage}[t]{.55\textwidth}
    \centering
      \caption{
      Comparison of different fitting strategies for FLD. 
      }
      \label{tab:trainedFLD}
      \footnotesize
      \setlength{\tabcolsep}{2.5pt}{
        \begin{tabular}{c|cc|cc}
          \toprule[2pt]
          \multirow{2}{*}{\textit{FLD Strategy}} & \multicolumn{2}{c|}{DanceTrack} & \multicolumn{2}{c}{SportsMOT} \\
           & HOTA & AssA & HOTA & AssA \\
          \midrule[1pt]
            -- & 51.1 & 33.4 & 73.7 & 61.5 \\
            Supervised & 51.4 & 34.0 & 77.6 & 68.0 \\
            History-Aware & \bf 57.7 & \bf 42.7 & \bf 81.1 & \bf 74.1 \\ 
          \bottomrule[2pt]
        \end{tabular}
        }
  \end{minipage}
\end{table}

\noindent \textbf{Supervised \textit{vs.} History-Aware.}
While our method achieves significant improvements over the baseline in \cref{tab:transformation}, these results alone do not fully substantiate our underlying motivation. A notable concern is whether this improvement stems from the introduction of FLD or the utilization of historical information as guidance, as we claimed.
To verify this, we fit an FLD using the ground-truth labels and features from the training set (denoted as \textit{supervised}) to serve as a competitor against our proposed approach.
Experiments in \cref{tab:trainedFLD} demonstrate that leveraging vast amounts of labeled data from different videos does not necessarily enable the FLD to identify a more appropriate projection space. This observation aligns with our argument in \cref{fig:sequence} and \cref{sec:introduction}, namely that the optimal subspace varies significantly across different sequences, simultaneously reinforcing that our underlying motivation holds enduring significance beyond the current implementation.
This insight may inspire future research to employ more advanced algorithmic designs to achieve this objective, rather than being limited to the use of FLD.

\noindent \textbf{History Length $T$.}
As stated at the end of \cref{sec:method-transformation}, we only consider ReID features from the $T$ most recent frames.
Although using a too-short temporal length $T$ decreases the credibility of the reference samples, it still provides a notable enhancement compared to the baseline tracker ($54.3~\textit{vs.}~51.1$ HOTA), as shown in \cref{tab:temporal-length}.
Conversely, an excessively large $T$ incorporates outdated features, making the distribution less representative and degrading performance.

\noindent \textbf{Temporally-Weighted Trajectory Centroid.}
Following our discussion in \cref{sec:method-temporal} about the varying temporal importance of features, we introduce the coefficient $\lambda_0$ to weight them accordingly.
\cref{tab:temporal-weight} demonstrates that using the temporal-weighted trajectory centroid can significantly enhance tracking performance. 
However, it is essential to note that excessively small values of $\lambda_0$ may lead to an overreliance on recent samples, resulting in a decline in robustness.

\noindent \textbf{Knowledge Integration.}
In \cref{tab:alpha}, we investigate various fusion coefficients $\alpha$ to balance robustness and specialization.
These results indicate that this involves a trade-off, prompting us to choose $0.9$ as our default setting.
In addition, this supports the concept outlined in \cref{sec:method-integration}, valuing the complementarity of those two spaces can boost the reliability of ReID features.

\section{Conclusion}
\label{sec:conclusion}

In this paper, we challenge a long-standing practice in multiple object tracking: \textbf{\textit{the direct adoption of appearance matching strategies from the re-identification task, an approach we argue is fundamentally inappropriate for tracking}}.
We contend that visual representations in MOT should be tailored to discriminate among the finite set in a given video sequence, as opposed to the open-set challenge.
Rather than the conventional approach of optimizing over training data, this should be an optimization specifically tailored to the current sequence that has been absent from previous methodologies.
To this end, we proposed an approach that leverages the tracking history to guide an adaptive transformation of the feature space, thereby boosting its discriminability.
Comprehensive experiments validate the effectiveness and versatility of our proposed approach and establish the new state-of-the-art performance.
These results provide compelling evidence that the potential of ReID features in MOT has been significantly underestimated, while simultaneously suggesting that our motivation holds enduring value for future explorations beyond this specific implementation.
Therefore, we hope our findings spur a wave of research into this crucial problem, whether in the form of new training-free components or as guiding principles for developing learnable modules.

\section*{Acknowledgements}
This work is supported by the Basic Research Program of Jiangsu (No. BK20250009), the Fundamental Research Funds for the Central Universities (No.020214380140), the Fundamental and Interdisciplinary Disciplines Breakthrough Plan of the Ministry
of Education of China (No. JYB2025XDXM118), the Collaborative Innovation Center of Novel Software Technology and Industrialization.

\appendix

\section{Experimental Details}
\label{sec:appendix-experimental-details}

\subsection{ReID Feature Discriminative Capability}
\label{sec:appendix-discriminative-capability}

As stated in \cref{sec:preliminary-analysis}, we adopt the metric $\delta_{1000}$ to quantify the discriminative capability of the representation space. This metric is derived from individual discriminative scores that are computed for each detection. 
Formally, for the $i$-th detection at time step $t$, we calculate the similarities against all history trajectories, as specified in \cref{eq:cost}. A discriminative score $\delta(t, i)$ for this detection is then defined as:

\begin{equation}
    \delta(t,i) = \text{Sim}^+(t, i, \tau_j) - \max_j \bigl[\text{Sim}^-(t, i, \tau_j)\bigr], 
    \label{eq:delta-1000}
\end{equation}
where $\text{Sim}^+(t, i, \tau_j)$ denotes the similarity to the corresponding positive sample (the $i$-th detection belongs to the $j$-th trajectory), and $\text{Sim}^-(t, i, \tau_j)$ denotes the similarity to a negative sample. We select the most similar negative sample using $\max_j \bigl[\text{Sim}^-(t, i, \tau_j)\bigr]$, because the most confusing example directly determines whether a misallocation of identities will occur.

After calculating all valid $\delta(t, i)$ within a video sequence, we aggregate them to obtain the overall discriminative measure.
Since tracking errors like ID switches occur in a very small portion of a long video (thousands of frames), we select the $1000$ most challenging cases from all discriminative scores.
In practice, we sort the scores in ascending order and select the smallest $1000$ samples to compute the averaging score $\delta_{1000}$, since these items are the most likely to be misassigned in tracking.

\subsection{ReID-based Tracker: FastReID-MOT}
\label{sec:appendix-reid-based-tracker}

As stated in \cref{sec:preliminary-tracker} and \cref{sec:experiments-implementation-details}, our baseline FastReID-MOT relies solely on ReID features for tracking. To keep the baseline straightforward, we implement a single-stage online tracker with a minimal set of hyperparameters:

\begin{itemize}
    \item $\lambda$, the feature update ratio in \cref{eq:ema}.
    \item $\theta_{\text{det}}$, detections with a confidence exceeding this threshold are considered by the tracker.
    \item $\theta_{\text{sim}}$, identity assignments with a similarity score exceeding this threshold are considered as valid choices.
    \item $\theta_{\text{new}}$, unmatched detections with a confidence exceeding this threshold are considered as newborn targets.
    \item $\theta_{\text{miss}}$, a trajectory is terminated if the number of consecutive missing frames is greater than this threshold.
\end{itemize}
All the aforementioned hyperparameters are tuned using a grid search on the corresponding datasets to maximize the baseline's performance.
In subsequent ablation experiments, we do not adjust these hyperparameters to ensure that the observed improvements are purely attributable to our proposed method.

\subsection{MASA Details}
\label{sec:appendix-masa}

In the MASA \cite{MASA} inference process, we simplified the original bi-softmax matching procedure \citep{MASA, QDTrack} to the simple cosine similarity combined with the Hungarian algorithm (as we detailed in \cref{sec:preliminary-tracker} and \cref{eq:cost}), and tuned some hyperparameters, which resulted in a slight improvement in tracking performance across all datasets. 
For our hyperparameters, we primarily adhered to the default settings outlined in \cref{sec:experiments-ablation}, with the exception of adjusting $\alpha$ to $0.5$ to better accommodate MASA's feature representation.

\subsection{Oracle Setting}
\label{sec:appendix-oracle}

In main paper \cref{tab:transformation} and \cref{fig:analysis-and-transformation-dancetrack}, we leverage an \textit{oracle setting} to focus our analysis on tracking performance without the influence of other factors.
In these experiments, we use the bounding boxes' coordinates from the ground truth files as the detection results and set all confidence scores to $1.0$. Even under these ideal conditions, the Detection Accuracy (DetA) will not reach $100.0$, as a result of the metric's calculation method \citep{HOTA}.

\subsection{Ablation Study}
\label{sec:appendix-ablation}

As we stated in \cref{sec:experiments-ablation}, the ablation experiments are conducted incrementally, with each table adding one component at a time:

\begin{itemize}
    \item In \cref{tab:transformation}, we apply $T=60$, $\lambda_0 = 1.0$ and $\alpha = 1.0$, which means we do not use the \textit{temporally-weighted trajectory centroid} and \textit{knowledge integration}.
    \item In \cref{tab:temporal-length}, we apply $\lambda_0 = 1.0$ and $\alpha = 1.0$, which means we do not use the \textit{temporally-weighted trajectory centroid} and \textit{knowledge integration}.
    \item In \cref{tab:temporal-weight}, we apply $T=60$ and $\alpha = 1.0$, which means we do not use the \textit{knowledge integration}.
    \item In \cref{tab:alpha}, we apply $T=60$ and $\lambda_0 = 0.9$, which means both proposed components are used in these experiments.
\end{itemize}
Together, these settings make up our default configuration ($T=60$, $\lambda_0=0.9$, $\alpha=0.9$) and are applied uniformly to all datasets as the default, as mentioned in \cref{sec:experiments-sota} and \cref{sec:experiments-ablation}.

\subsection{Visualization of ReID features}
\label{sec:appendix-visualization-details}
To qualitatively evaluate the discriminative capability of ReID features, we visualize feature similarities both within and across sequences. In \cref{fig:space-one-sequence}, we show the features of objects in $15$ consecutive frames of a single video sequence, projected to a two-dimensional space using Principal Component Analysis (PCA). In \cref{fig:space-all-sequence}, we randomly select $10$ sequences from the DanceTrack dataset \citep{DanceTrack} and visualize features extracted from $40$ consecutive frames of each sequence, also projected via PCA.

\section{More Results}
\label{sec:appendix-results}

\begin{table}[tb]
  \centering
  \caption{
  Performance comparison with state-of-the-art methods on MOT17 \citep{MOT16}.
  The best and second-best results are denoted in \textbf{bold} and \underline{underline}, respectively.
  }
  \small
  \setlength{\tabcolsep}{4pt}{
    \begin{tabular}{l|ccccc}
      \toprule[2pt]
      Methods & HOTA & DetA & AssA & IDF1 \\
      \midrule[1pt]
      \textit{\underline{motion-based:}} \\
      ByteTrack \citep{ByteTrack} & 63.1 & 64.5 & 62.0 & 77.3 \\
      OC-SORT \citep{OC-SORT} & 63.2 & 63.2 & 63.4 & 77.5 \\
      C-BIoU \citep{C-BIoU} & 64.1 & 64.8 & 63.7 & 79.7 \\
      \midrule[0.5pt]
      \textit{\underline{reid-based:}} \\
      QDTrack \citep{QDTrack} & 53.9 & 55.6 & 52.7 & 66.3 \\
      ContrasTR \citep{ConstrastiveDETR-MOT} & 58.9 & -- & -- & 71.8 \\
      FastReID-MOT \small (baseline) & \underline{61.5} & \underline{63.4} & \underline{60.0} & \underline{73.5} \\
      HAT-FastReID-MOT$\dagger$ & \bf 63.5 & \bf 64.0 & \bf 63.2 & \bf 77.5 \\
      \midrule[0.5pt]
      \textit{\underline{hybrid-based:}} \\
      FairMOT \citep{FairMOT} & 59.3 & 60.9 & 58.0 & 72.3 \\
      DeepSORT \citep{Deep-SORT} & 61.2 & 63.1 & 59.7 & 74.5 \\
      MixSort-OC \citep{SportsMOT} & 63.4 & 63.8 & 63.2 & 77.8 \\
      DiffMOT \citep{DiffMOT} & \bf 64.5 & \bf 64.7 & \bf 64.6 & \bf 79.3 \\
      OC-SORT-ReID & 64.1 & \underline{64.4} & 64.0 & 79.0 \\
      HAT-OC-SORT-ReID & \underline{64.2} & \underline{64.4} & \underline{64.1} & \underline{79.2} \\
      \bottomrule[2pt]
    \end{tabular}
  }
  \label{tab:appendix-sota-mot17}
\end{table}

\begin{figure}[tb]
    \centering
    \includegraphics[width=0.6\linewidth]{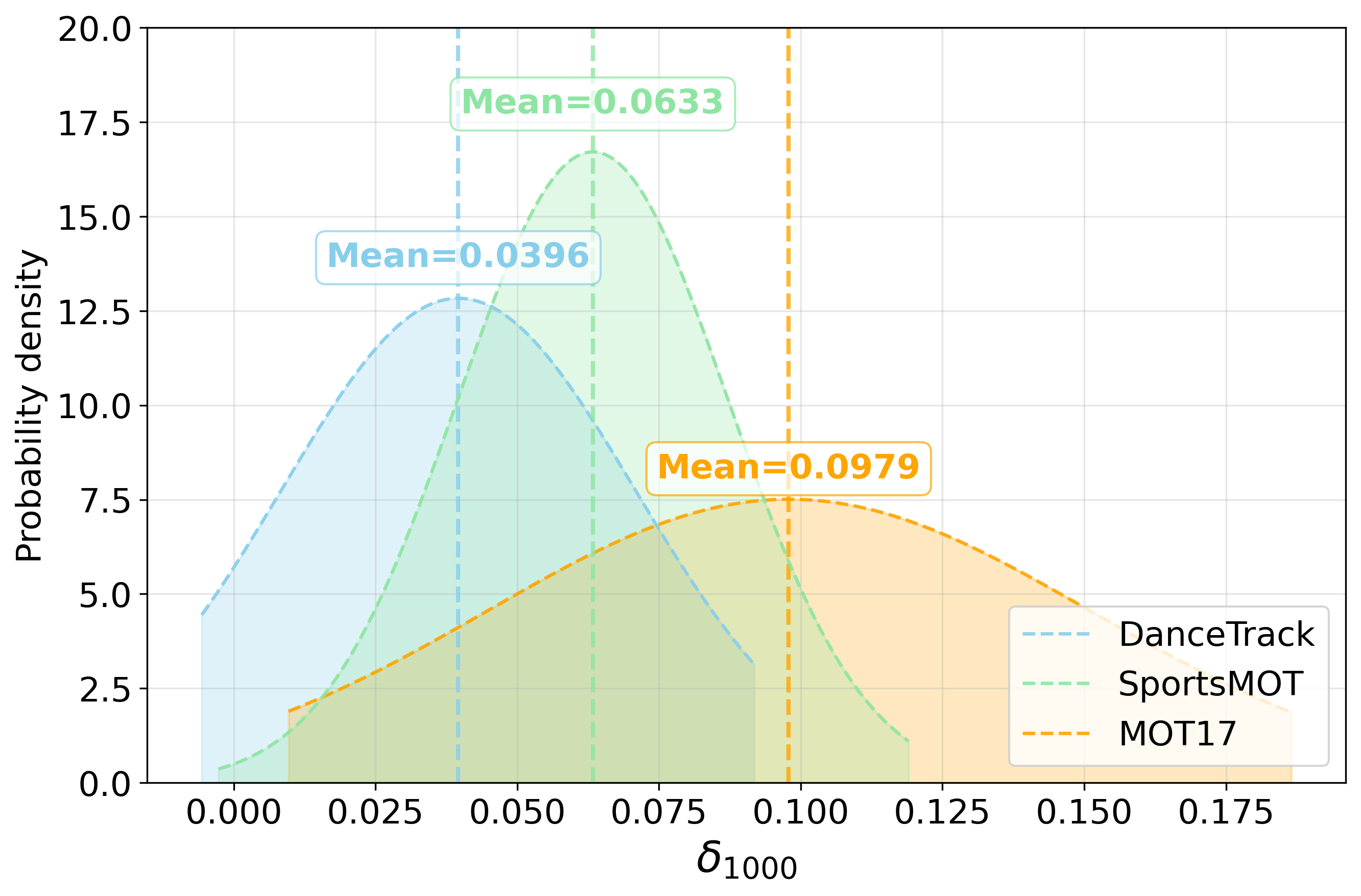}
    \caption{
    Comparison of ReID separability on DanceTrack \citep{DanceTrack}, SportsMOT \citep{SportsMOT}, and MOT17 \citep{MOT16} based on $\delta_{1000}$.
    }
    \label{fig:appendix-mot17-distribution}
\end{figure}

\subsection{MOT17}
\label{sec:appendix-mot17}

In \cref{tab:appendix-sota-mot17}, we present our experimental results on the MOT17 \citep{MOT16} dataset. 
Due to the submission limits of the MOT17 evaluation server, we built our hybrid-based tracking using only the classic OC-SORT \citep{OC-SORT} algorithm.
Compared to our baseline (FastReID-MOT), our method yields a significant performance gain ($2.0$ HOTA and $3.2$ AssA), though the margin is not as large as on other benchmarks \citep{DanceTrack, SportsMOT}. We attribute this to the fact that the MOT17 dataset, consisting solely of pedestrians, has high inherent target discriminability (\textit{e.g.}, distinct clothing colors and styles), which limits the room for our method to make a greater impact.
In the hybrid-based experiments, we do not achieve a highly satisfactory performance. 
On the one hand, prior studies \citep{ByteTrack, C-BIoU} have shown that the simple motion patterns within MOT17 allow the motion prediction module to take a dominant role, thereby constraining the influence of the ReID branch. Our observations in \cref{fig:appendix-mot17-distribution}, based on $\delta_{1000}$, also confirm this. The ReID features of MOT17 targets show significantly greater separability, despite the dataset containing up to ten times more targets per frame compared to DanceTrack \citep{DanceTrack} and SportsMOT \citep{SportsMOT}.
On the other hand, the overly engineered fusion of multiple modules and the unreliable validation set split further increased the difficulty of optimizing the entire method.
Despite these challenges, we still outperform MixSort-OC \citep{SportsMOT} that also uses OC-SORT as the framework, and are slightly behind \citet{DiffMOT}, which is based on learnable motion estimation.

To summarize, although our method does not achieve flawless results on MOT17, the consistent performance gains across experiments robustly demonstrate its effectiveness and applicability in diverse scenarios. 
Coupled with its outstanding performance across various other scenarios \citep{DanceTrack, SportsMOT, TAO} in \cref{tab:sota-dancetrack}, \cref{tab:sota-sportsmot} and \cref{tab:sota-masa}, our method still holds enough promise and is attractive for future exploration.

\subsection{MOT20}
\label{sec:appendix-mot20}

\begin{table}[tb]
  \centering
  \caption{
  Performance comparison with state-of-the-art methods on the MOT20 validation set.
  The best results are denoted in \textbf{bold}.
  }
  \label{tab:sota-mot20-val}
  \small
  \setlength{\tabcolsep}{4pt}{
    \begin{tabular}{l|ccccc}
      \toprule[2pt]
      Methods & HOTA & DetA & AssA & MOTA & IDF1 \\
      \midrule[1pt]
      Deep OC-SORT \tiny \citep{Deep-OC-SORT} & 59.5 & -- & 58.2 & -- & 76.3 \\
      Hybrid-SORT-ReID \tiny \citep{Hybrid-SORT}
      & 60.7 & 61.6 & 60.0 & 74.0 & 78.4 \\
      FastReID-MOT 
      & 57.7 & 61.7 & 54.1 & 74.5 & 72.4 \\ 
      HAT-FastReID-MOT 
      & \bf 61.2 & \bf 62.3 & \bf 60.4 & \bf 75.0 & \bf 78.8 \\
      \bottomrule[2pt]
    \end{tabular}
  }
\end{table}

We also evaluate our method on MOT20 \citep{MOT20}. To ensure fairness, all algorithms are implemented using the same public FastReID \citep{FastReID} weight. 
Due to current technical issues with the official MOTChallenge evaluation server that prevent successful submissions to the MOT20 test set\footnote{At the time of submission, the online evaluation on Codabench consistently terminates with a ``Soft time limit exceeded'' error. Related discussions can be found at \url{https://www.codabench.org/forums/9893/1624/}.}, we report our comparative results on the validation set. As reported in \cref{tab:sota-mot20-val}, our proposed method consistently outperforms both advanced trackers and our FastReID-MOT baseline. These results demonstrate the effectiveness of the proposed history-aware feature transformation under crowded scenes, and further validate the generalization ability of our method.

\begin{table}[p]
  \centering     
  \begin{adjustbox}{angle=90, center}
  \begin{minipage}[t]{1.2\textwidth}
    \centering
    \centering
\caption{
Detailed performance comparison with state-of-the-art methods on the Dancetrack test set.
By default, higher values indicate better performance, while metrics marked with $\downarrow$ denote that lower values are better.
}
\label{tab:sota-dancetrack-detail}
\small
\setlength{\tabcolsep}{1.4pt}{
\begin{tabular}{l|ccccccccccc}
\toprule[2pt]
Methods & HOTA & DetA & AssA & LocA & MOTA & IDF1 & IDR & IDP & IDTP & IDFN$\downarrow$ & IDFP$\downarrow$ \\
\midrule[1pt]

\textit{\underline{motion-based:}} \\
ByteTrack \tiny \citep{ByteTrack} 
& 47.7 & 71.0 & 32.1 & - & 89.6 & 53.9 & - & - & - & - & - \\
DiffusionTrack \tiny \citep{DiffusionTrack} 
& 52.4 & 82.2 & 33.5 & - & 89.3 & 47.5 & - & - & - & - & - \\
OC-SORT \tiny \citep{OC-SORT} 
& 55.1 & 80.3 & 38.3 & - & 92.0 & 54.6 & - & - & - & - & - \\
C-BIoU \tiny \citep{C-BIoU} 
& 60.6 & 81.3 & 45.4 & - & 91.6 & 61.6 & - & - & - & - & - \\

\midrule[0.5pt]
\textit{\underline{ReID-based:}} \\
QDTrack \tiny \citep{QDTrack} 
& 54.2 & 80.1 & 36.8 & - & 87.7 & 50.4 & - & - & - & - & - \\
FastReID-MOT \tiny (our baseline)
& 50.6 & 81.1 & 31.6 & 92.5 & 90.3 & 50.4 & 48.6 & 52.4 & 140635 & 148531 & 127941 \\

HAT-FastReID-MOT
& 58.6 & 81.3 & 42.3 & 92.6 & 89.6 & 57.9 & 55.7 & 60.4 & 161074 & 128092 & 105783 \\

HAT-FastReID-MOT$\dagger$
& \bf 61.2 & \bf 81.6 & \bf 46.0 & \bf 92.7 & \bf 89.7 & \bf 61.1 & \bf 58.7 & \bf 63.7 & \bf 169663 & \bf 119503 & \bf 96884 \\

\midrule[0.5pt]
\textit{\underline{hybrid-based:}} \\
FairMOT \tiny \citep{FairMOT} 
& 39.7 & 66.7 & 23.8 & - & 82.2 & 40.8 & - & - & - & - & - \\
DeepSORT \tiny \citep{Deep-SORT} 
& 45.6 & 71.0 & 29.7 & - & 87.8 & 47.9 & - & - & - & - & - \\
StrongSORT \tiny \citep{StrongSORT} 
& 55.6 & 80.7 & 38.6 & - & 91.1 & 55.2 & - & - & - & - & - \\
DiffMOT \tiny \citep{DiffMOT} 
& 62.3 & \bf 82.5 & 47.2 & - & \bf 92.8 & 63.0 & - & - & - & - & - \\
Hybrid-SORT-ReID \tiny \citep{Hybrid-SORT} 
& 65.7 & - & - & - & 91.8 & 67.4 & - & - & - & - & - \\
ByteTrack-ReID 
& 52.4 & 71.0 & 38.7 & 85.1 & 87.9 & 60.4 & 58.2 & 62.7 & 168175 & 120991 & 99897 \\

HAT-ByteTrack-ReID 
& 56.1 & 71.4 & 44.2 & 85.1 & 88.5 & 65.7 & 63.6 & 68.0 & 183838 & 105328 & 86371 \\

OC-SORT-ReID 
& 60.8 & 81.0 & 45.7 & 92.4 & 90.6 & 63.5 & 61.2 & 65.9 & 177073 & 112093 & 91589 \\

HAT-OC-SORT-ReID 
& 64.6 & 81.5 & 51.3 & \bf 92.6 & 90.3 & 67.7 & 65.1 & 70.4 & 188348 & 100818 & 79266 \\

HAT-Hybrid-SORT-ReID 
& \bf 66.9 & 81.5 & \bf 55.0 & \bf 92.6 & 90.5 & \bf 71.3 & \bf 68.7 & \bf 74.2 & \bf 198722 & \bf 90444 & \bf 69211 \\

\bottomrule[2pt]
\end{tabular}
}

  \end{minipage}
  \end{adjustbox}
\end{table}

\subsection{Extended Evaluation with Additional Metrics}
\label{appendix-extended-metrics}

To provide a more comprehensive and fine-grained evaluation of tracking performance, we report an extended set of metrics in \cref{tab:sota-dancetrack-detail}. These complementary metrics allow a more thorough assessment of detection accuracy, association robustness, and identity consistency.

\subsection{Comparison with end-to-end methods}
\label{appendix-compare-with-E2E-methods}


End-to-end (E2E) trackers and heuristic tracking-by-detection methods follow fundamentally different paradigms, which makes direct comparisons inherently unfair and scenario-dependent. To provide a complete perspective, we nevertheless report comparisons with representative E2E methods on both DanceTrack and SportsMOT.

On DanceTrack, our method achieves a HOTA score of $66.9$, which is competitive with recent E2E approaches, exceeding CO-MOT\citep{CO-MOT} and being comparable to SambaMOTR\citep{SambaMOTR}. On SportsMOT, our method significantly outperforms all existing published E2E trackers, e.g., $81.2$ versus $72.6$ of MOTIP\citep{MOTIP}, demonstrating a clear advantage. These results indicate that neither E2E nor heuristic methods uniformly dominate across all datasets. Instead, their relative effectiveness is highly scenario-dependent. Our method demonstrates strong competitiveness against state-of-the-art E2E models on DanceTrack and achieves decisive superiority on SportsMOT, further validating the practical value and versatility of the proposed framework.

\subsection{Inference Speed}
\label{sec:appendix-inference-speed}

Given the detection results (without the latency of detectors), our method (including the ReID model \citep{FastReID}) achieves an inference speed of $22.7$ FPS, compared to $46.5$ FPS for the baseline, on DanceTrack \citep{DanceTrack} using an NVIDIA RTX A5000 GPU and an AMD Ryzen 9 5900X CPU.
Although this meets the requirements for near real-time tracking, we must point out two main challenges that remain for achieving faster inference.

Based on our experiments, nearly all of the additional latency originates from the computation of eigenvalues and eigenvectors, as this operation runs on the CPU (with \texttt{scipy.linalg.eigh(S\_B, S\_W)}), which is inherently inefficient for matrix calculations.
We explored some alternative GPU-based packages like PyTorch, JAX, and CuPy. These packages offer CUDA acceleration for eigenvector computations (\texttt{eigh()} function).
However, they lack an interface for generalized eigenvalue solving in \texttt{eigh()} (\textit{e.g.}, discussed in \#5461 issue\footnote{\url{https://github.com/jax-ml/jax/issues/5461}} in the official repository of JAX, it only accepts one matrix for the eigenvalue calculation), which is a feature provided by SciPy and used for FLD solution.
Transforming the input into a format acceptable for these functions incurs additional computational overhead and results in a loss of precision.
If the same interface can be used, we estimate, based on experience, that it would result in a $4\times$ to $10\times$ speedup.

Moreover, the redundancy in feature dimensions further exacerbates this issue ($2048$ from FastReID \citep{FastReID} \textit{vs.} $256$ from MASA \citep{MASA}), since latency increases with dimension count.
This issue could be mitigated by either employing other dimensionality reduction methods or by reducing the output dimension of the ReID feature head during the training phase.
In addition, some computational procedures are performed trajectory by trajectory (like centroid computation), which can slow down inference as target density increases \cite{MOT16}. This also suggests a direction for future implementation optimization.

In summary, we consider that addressing this operator issue falls beyond the scope of this paper as it pertains to a complicated engineering problem.

\subsection{Visualizations}
\label{sec:appendix-visualizations}

\begin{figure}[tb] 
    \centering 
    \begin{minipage}{0.48\textwidth} 
        \centering 
        \includegraphics[width=0.95\linewidth]{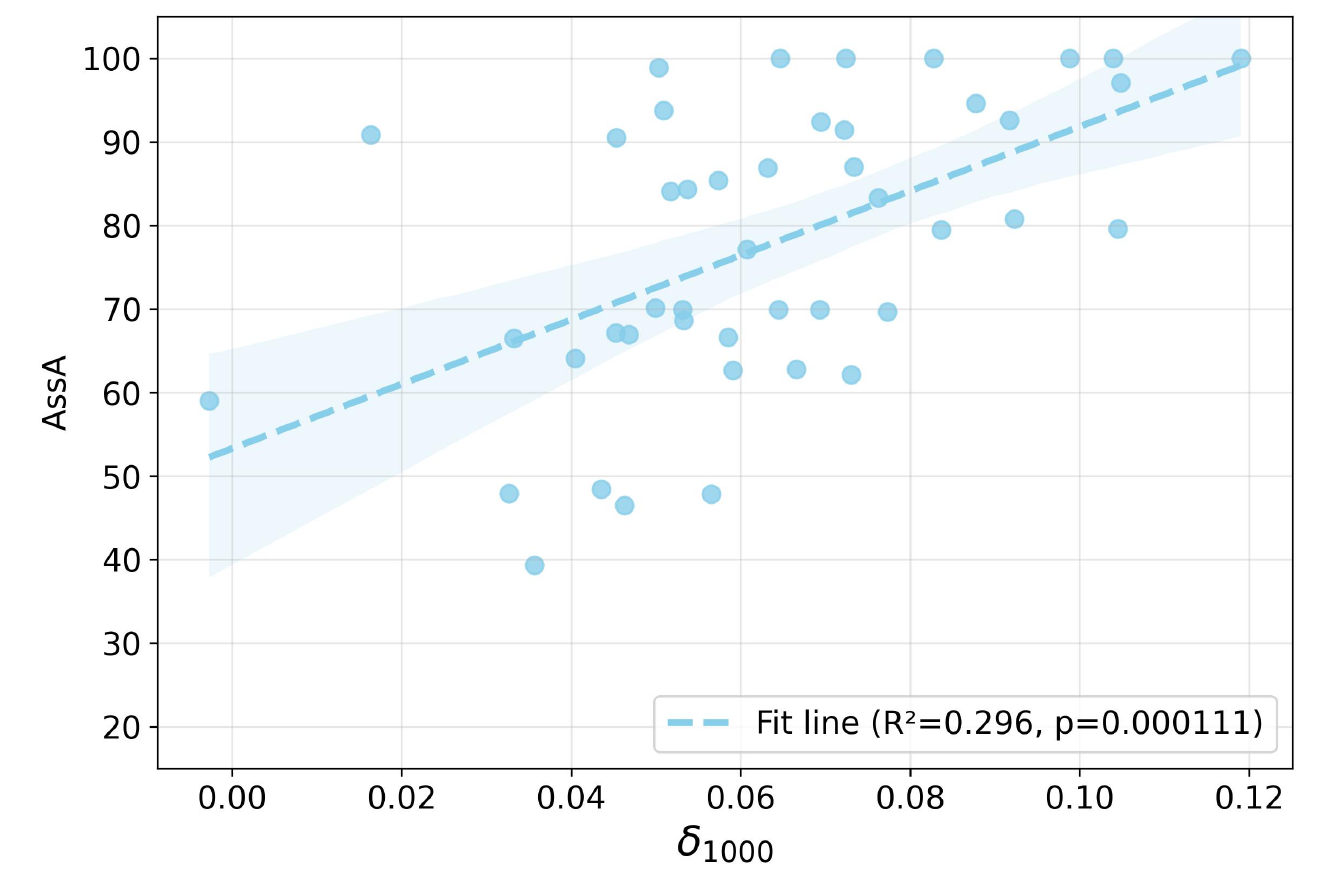} 
        \subcaption{Significant and reliable positive correlation between discriminability and tracking performance.}
        \label{fig:analysis-sportsmot}
    \end{minipage}
    \hfill 
    \begin{minipage}{0.48\textwidth} 
        \centering
        \includegraphics[width=0.95\linewidth]{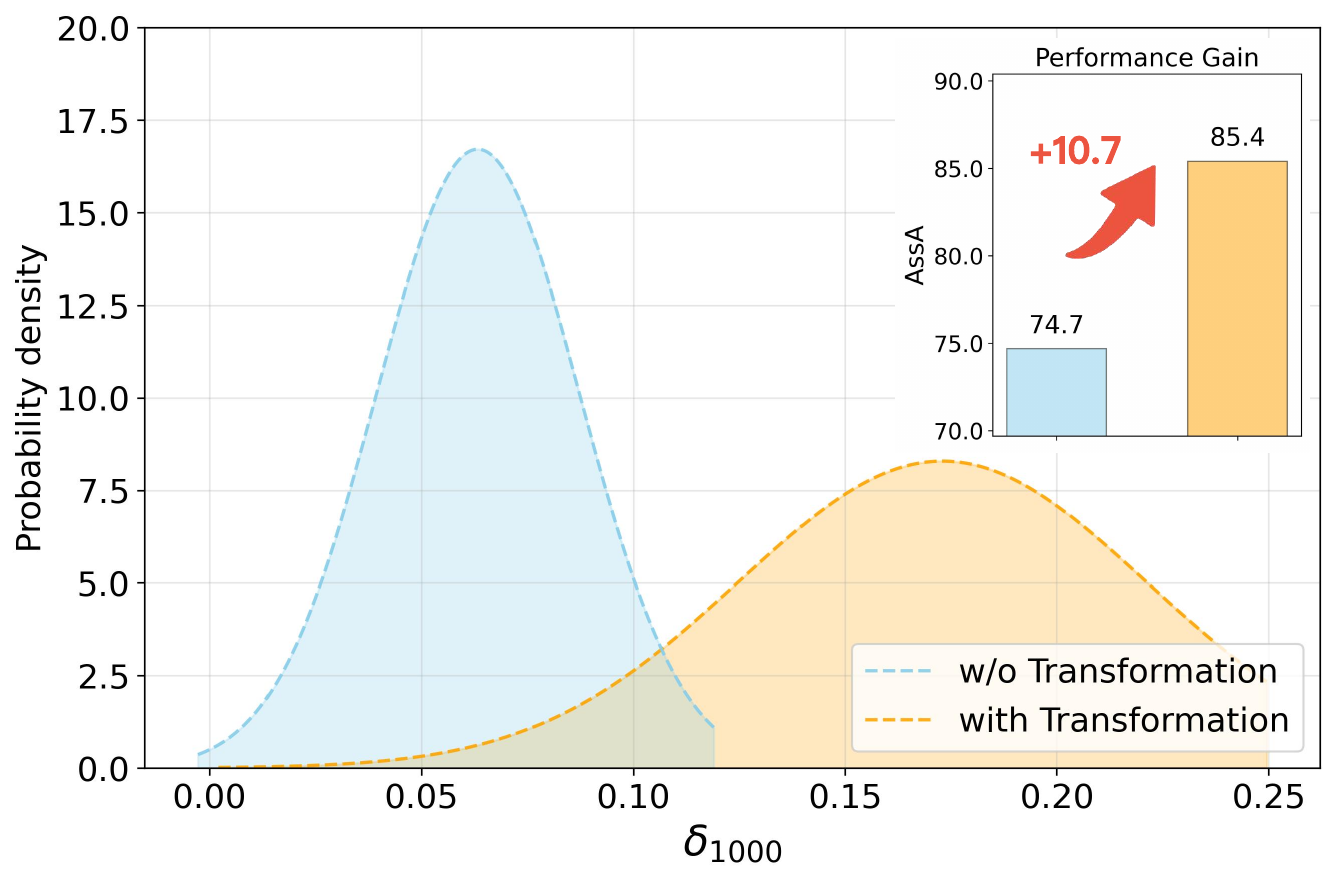}
        \subcaption{Our transformation improves performance by enhancing feature discriminative capability.}
        \label{fig:transformation-sportsmot}
    \end{minipage}
    \caption{Correlation between ReID feature discriminability $\delta_{1000}$ and tracking accuracy AssA on SportsMOT \citep{SportsMOT}.}
    \label{fig:analysis-and-transformation-sportsmot}
\end{figure}

\subsubsection{Discriminative Capability Analysis on SportsMOT}
\label{sec:appendix-analysis-sportsmot}

As stated in \cref{sec:preliminary-analysis}, we observe a significant and reliable positive correlation between the discriminative capability ($\delta_{1000}$) of ReID features and the object association accuracy (AssA~\citep{HOTA}) on DanceTrack \citep{DanceTrack}. 
To further examine the generality of this relationship, we extend the analysis to the SportsMOT dataset \citep{SportsMOT}. 
As shown in \cref{fig:analysis-sportsmot}, the visualizations on SportsMOT also demonstrate a consistently positive and statistically meaningful correlation between $\delta_{1000}$ and AssA, in agreement with the findings on DanceTrack in \cref{fig:analysis-dancetrack}. 
It strongly supports our direction: improving discriminative capability to boost tracking performance.

\cref{fig:transformation-sportsmot} validates that our transformation can enhance feature discriminability to improve tracking performance on SportsMOT \citep{SportsMOT}. This result, echoing the findings in \cref{fig:transformation-dancetrack}, further substantiates our core hypothesis.

\subsection{Visualization of ReID Features}

\begin{figure}[tb]
    \centering
    \includegraphics[width=0.9\linewidth]{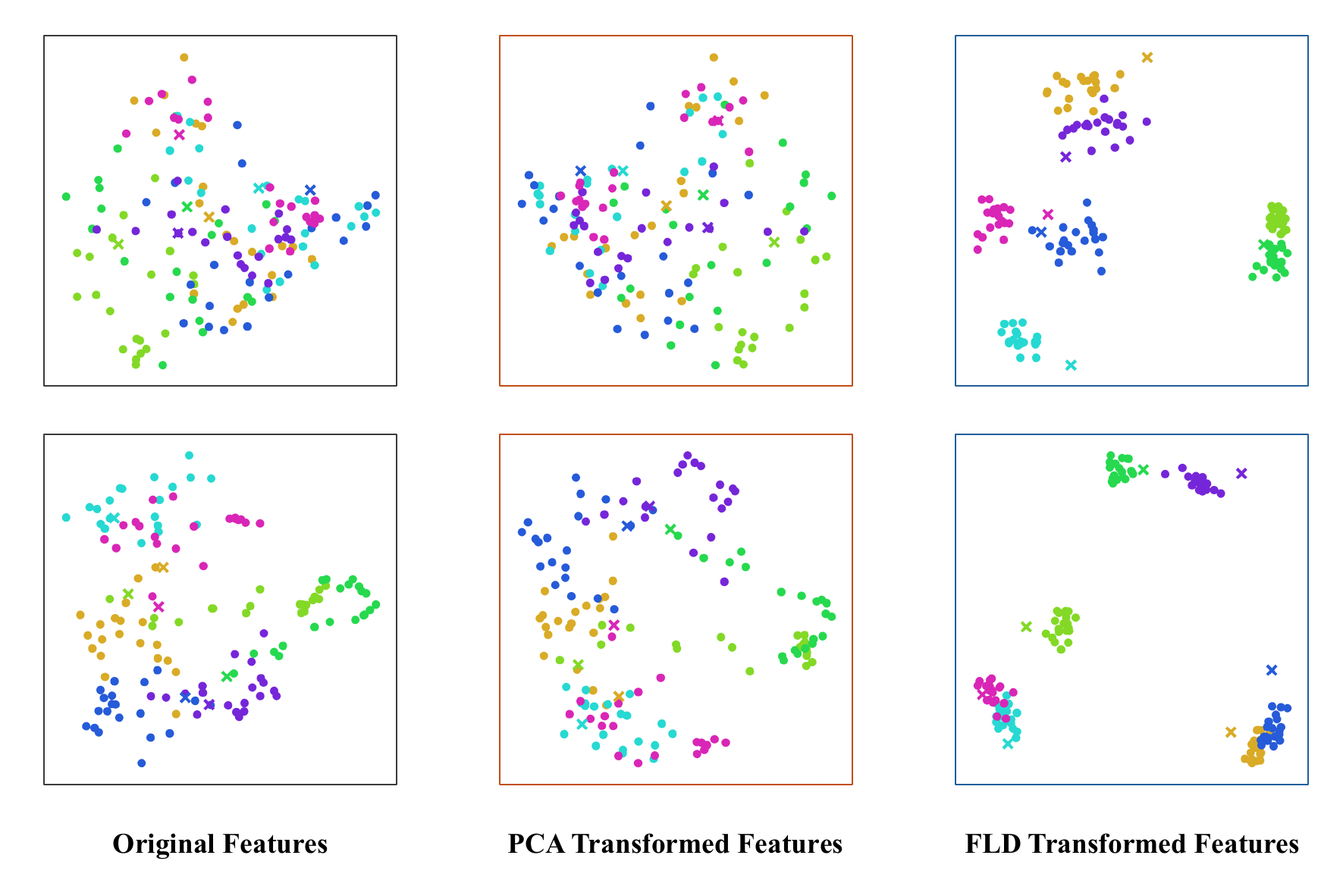}
    \caption{
    \textbf{Visualization of ReID features.}
    $\bullet$ represents the historical features and \faTimes ~indicates the current features.
    Compared to the other two spaces, the FLD-projected space shows better differentiation of trajectories.
    }
\label{fig:appendix-more-projections}
\end{figure}

To further assess the impact of feature transformation on ReID discriminability, we visualize the features in different linear projection spaces in \cref{fig:appendix-more-projections}.
Features transformed by FLD exhibit clearer inter-trajectory separation than those produced by PCA or the original space. Taken together, the quantitative gains reported in main paper \cref{tab:transformation} and the qualitative improvements observed in the visualizations indicate that incorporating historical trajectory information into the projection step is a principled and effective strategy for improving multiple object tracking: historical trajectories constitute an invaluable supervisory signal for representation selection and should therefore be exploited in the reasoning pipeline rather than disregarded.

\section{Limitations}
\label{sec:appendix-limitations}

While our method has yielded encouraging results, there are some limitations and concerns that need to be pointed out.

\noindent \textbf{Hybrid-based Tracker.} 
While our method demonstrates significant improvements for ReID-based trackers, its gains on hybrid-based methods are somewhat limited. Besides the saturated metrics and overly complex algorithmic design discussed in Sec. 5.3, a deeper, more fundamental bias lies at the core: current hybrid-based trackers prioritize location information.
For instance, in existing hybrid-based methods \citep{Deep-OC-SORT, Hybrid-SORT, DiffMOT}, the assignment stage often relies entirely or heavily on the IoU metric. This leads to the ReID information being either overlooked or not sufficiently trusted, thereby creating a disconnect between the ReID branch and performance improvement. 
Our method enhances the trustworthiness of ReID features, which may inspire future hybrid-based methods to develop ReID-first or more ReID-reliant trackers. We believe this could significantly alter the algorithmic logic of existing trackers, which we leave for future work to explore.

\noindent \textbf{End-to-End Method.} 
A potential concern is that our method cannot be applied to state-of-the-art end-to-end models \citep{SambaMOTR, CO-MOT, MOTIP}.
First, we argue that heuristic and end-to-end methods represent two distinct paths to the same goal, with no inherent superiority of one over the other, a common phenomenon in computer vision \citep{DETR, YOLOX, DBLP:conf/nips/DhariwalN21, DBLP:journals/corr/abs-2406-06525}. Therefore, our proposed method does not need to compete directly with end-to-end approaches, and its inability to serve them is acceptable. This does not diminish the value of our method.
Second, while our proposed history-aware transformation cannot be directly applied to end-to-end methods (\textit{e.g.}, track queries), we believe it offers a valuable philosophical insight. Specifically, the observation that the information disparity between intra- and inter-trajectory features in historical tracklets can help a model better distinguish different tracks and thus improve tracking performance.
This insightful conclusion might help guide the design of trainable or end-to-end models, which could potentially enable our ideas to extend beyond the realm of heuristic algorithms.

\section{Discussions}
\label{sec:appendix-discussions}

\noindent
\textbf{Hyperparameter Sensitivity.}
Although we conduct extensive exploratory experiments to select hyperparameters, the performance gains brought by our method are not sensitive to them.
As shown in \cref{tab:temporal-length} - \cref{tab:alpha}, although the hyperparameters introduce performance fluctuations, the positive gains remain consistent.
The results without $\dag$ in \cref{tab:sota-dancetrack} and \cref{tab:sota-sportsmot} use the ablation-selected settings from DanceTrack and are directly transferred to other datasets without further fine-tuning.
In this paper, hyperparameter selection is more of an engineering choice aimed at achieving the best metrics.

\noindent
\textbf{Comparison with More Recent Methods.}
Our comparison mainly focuses on our proposed baseline and several trackers \cite{OC-SORT, Hybrid-SORT} published about two years ago.
We did not include some more recent trackers for two main reasons. First, most recent hybrid trackers rely on additional cues for enhancement, which can dilute the final gains from ReID-only improvements. Second, many high-performing heuristic algorithms are highly dependent on hyperparameter tuning for their final metrics, which increases the difficulty of iterative exploration.
This work aims more to reveal a potentially long-overlooked issue through reliable systematic analysis than to pursue absolute state-of-the-art performance. We hope to devote more effort to systematic exploration rather than experimental engineering iterations. 
This may point to a potential limitation and an important consideration for future work or real-world applications.

\noindent
\textbf{Unsuccessful Technical Attempts.}
Beyond the customized FLD strategies ultimately adopted in this work, we also explored several techniques that did not yield effective gains during our research.
On the one hand, we considered incorporating detection confidence into the Weighted Trajectory Centroid to adjust the centroid position together with temporal information and make it more reliable, but this did not yield significant gains.
On the other hand, we also explored using discriminability (like cosine similarity) to adjust the contribution of inter-trajectory similarities to the final solution, with the goal of encouraging the projection matrix to focus more on hard-to-distinguish trajectories. However, this also did not yield consistent gains.
The conclusions presented here are not definitive, but rather empirical observations from our exploration, intended to inform potential future research.

%
%
\bibliographystyle{splncs04}
\bibliography{main}

\end{document}